\documentclass[10pt,twocolumn,letterpaper]{article}

\usepackage{iccv}
\usepackage{times}
\usepackage{epsfig}
\usepackage{graphicx}
\usepackage{amsmath}
\usepackage{amssymb}
\usepackage{bbm}

\makeatletter
\@ifundefined{showcaptionsetup}{}{%
	\PassOptionsToPackage{caption=false}{subfig}}
\usepackage{subfig}
\makeatother
\usepackage[pagebackref=true,breaklinks=true,letterpaper=true,colorlinks,bookmarks=false]{hyperref}
\iccvfinalcopy % *** Uncomment this line for the final submission

\def\iccvPaperID{****} % *** Enter the ICCV Paper ID here

\ificcvfinal\fi

\begin{document}
%%%%%%%%% TITLE
\title{Wavelet-based Reflection Symmetry Detection via Textural and Color Histograms}
\author{Mohamed Elawady, Christophe Ducottet, Olivier Alata, C\'{e}cile Barat\\
Universit\'{e} Jean Monnet, CNRS, UMR 5516\\
Laboratoire Hubert Curien, F-42000, Saint-Etienne, France\\
{\small mohamed.elawady@univ-st-etienne.fr}
\and
Philippe Colantoni\\
Universit\'{e} Jean Monnet, CIEREC EA n\textsuperscript{0} 3068\\
Saint-\'{E}tienne, France
}

\maketitle
%\thispagestyle{empty}
%%%%%%%%%%%%%%%%%%%%%%%%%%%%%%%%%%%%%%%%%%%%%%%%%%%%%%%%%%%%%%%%%%%%%%%%%%%%%%%%%
\begin{abstract}
%- MAX : 4000 Chars
Symmetry is one of the significant visual properties inside an image plane, to identify the geometrically balanced structures through real-world objects. Existing symmetry detection methods rely on descriptors of the local image features and their neighborhood behavior, resulting incomplete symmetrical axis candidates to discover the mirror similarities on a global scale. In this paper, we propose a new reflection symmetry detection scheme, based on a reliable edge-based feature extraction using Log-Gabor filters, plus an efficient voting scheme parameterized by their corresponding textural and color neighborhood information. Experimental evaluation on four single-case and three multiple-case symmetry detection datasets validates the superior achievement of the proposed work to find global symmetries inside an image. 
\end{abstract}
%%%%%%%%%%%%%%%%%%%%%%%%%%%%%%%%%%%%%%%%%%%%%%%%%%%%%%%%%%%%%%%%%%%%%%%%%%%%%%%%%
\section{Introduction}
%-- Figure: Proposed framework
%-- What is global symmetry
%-- Paper focus
%-- Problem definition
Reflection symmetry is a geometrical property in natural and man-made scenes which attracts the human attention by achieving the state of equilibrium through the appearance of the similar visual patterns around \cite{Arnheim2001,Zakia2010}. It can be defined as a balanced region where local patterns are nearly matching on each side of a symmetry axis. This paper focuses on detecting the global mirror symmetries inside an image plane by highlighting the inter-correlation between the edge, textural and color information of major involved objects.   

Many detection algorithms of reflection symmetry has been introduced in the last decade. Loy and Eklundh \cite{Loy2006} proposed the baseline algorithm by analyzing the bilateral symmetry from image features' constellation. Firstly, local feature points (i.e. SIFT) are extracted associated with local geometrical properties and descriptor vectors. Secondly, the process of pairwise matching is computed based on the measure of the local symmetry magnitude of their descriptors. Finally, A Hough-like voting space is constructed, respect to the accumulation of symmetry candidates' magnitude, parametrized with orientation and displacement components, to select the dominant axes inside an image. Park et al. \cite{Park2008} presented a survey for symmetry detection algorithms, followed by two symmetry detection challenges \cite{Rauschert2011,Liu2013}, in which Loy and Eklundh \cite{Loy2006} have the best symmetrical results against the participated methods \cite{Mo2011,Kondra2013,Michaelsen2013,Patraucean2013}. Later, other feature-based approaches \cite{Cho2009,Cai2014} introduced some improvements over the baseline algorithm \cite{Loy2006}. Edge-based features \cite{Ming2013,Cicconet2014,Wang2015,Atadjanov2015,Cicconet2016,Atadjanov2016,Elawady2016} were more recently proposed instead of the feature points, due to their robust behavior in detecting an accurate symmetric information inside an image plane.
\begin{figure}[]
    \centering
    \includegraphics[width=0.9\columnwidth]{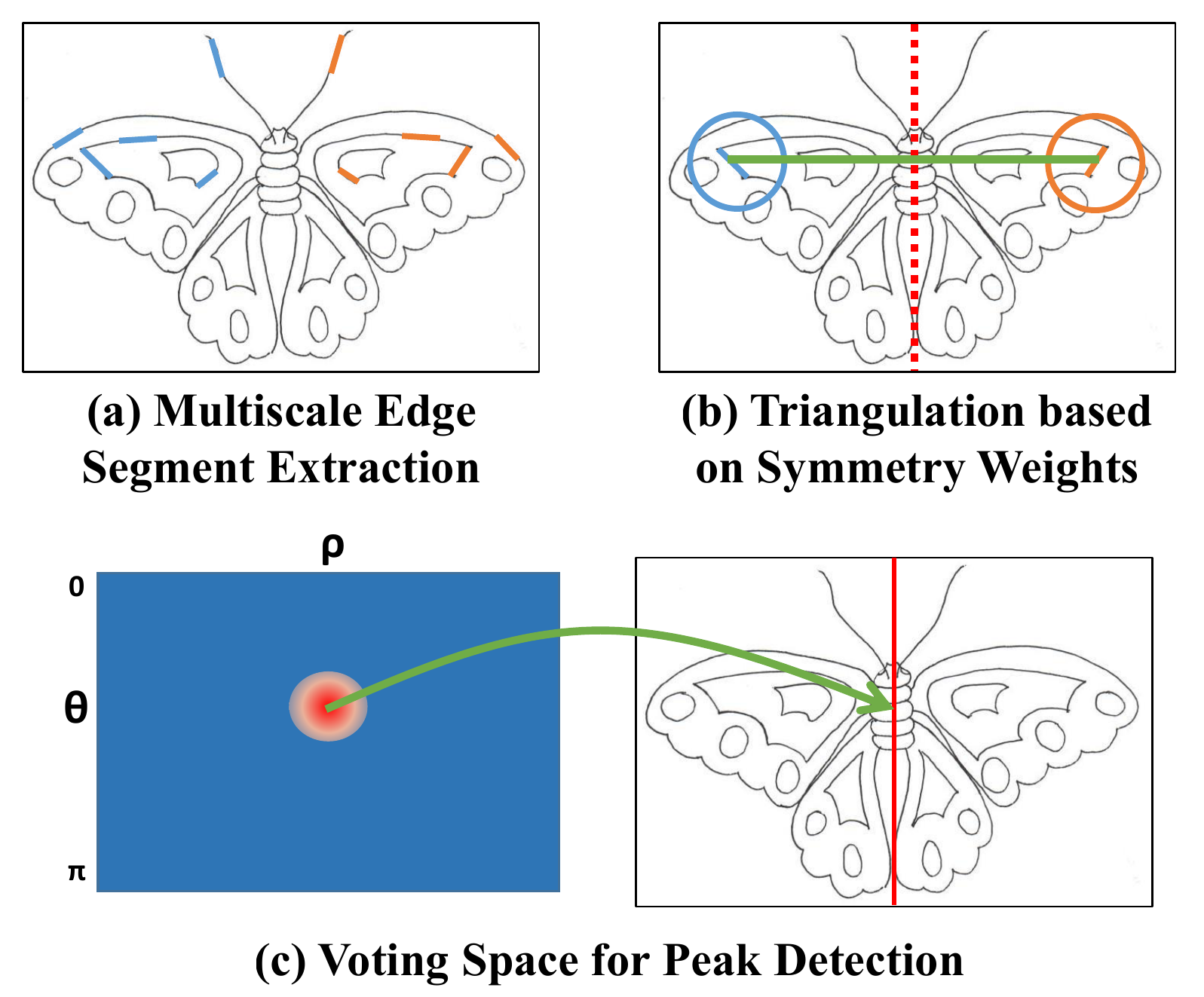}
    \caption{The general framework of the proposed reflection symmetry detection, using multi-scale edge detection followed by symmetry triangulation based on local geometrical, spatial and color information.}
    \label{fig:framework}
\end{figure}

Log-Gabor filter was introduced by Field \cite{Field1987} in late 1980's, as an alternative to the Gabor filter (formerly used in \cite{Elawady2016}), to suppress the effect of DC component through the computation of the multi-scale logarithmic function in the frequency domain. This process extracts more accurate feature (edge/corner) information from an image, plus ensuring the robustness of these features with respect to illumination variations. Feature extraction methods based on Log-Gabor filter have been successfully used in different computer vision applications (biometric authentication \cite{kaur2017cancelable,ahmed2017combining}, image retrieval \cite{walia2016boosting}, face recognition \cite{li2015log}, image enhancement \cite{Wang2008}, character segmentation \cite{Mancas-Thillou2006}, and edge detection \cite{Gao2007}).

Our contribution is twofold. Firstly, we extract edge features upon the application of Log-Gabor filters for symmetry detection instead of using Gabor filters. Secondly, we propose a similarity measure based on color image information, to improve the symmetry magnitude estimation in the voting step. In addition, we evaluate the proposed methods over all public datasets for reflection symmetry (single and multiple cases), in comparison with state-of-the-art algorithms.

Figure~\ref{fig:framework} shows our proposed algorithm, which detects globally the symmetry axes inside an image plane. We firstly extract edge features using Log-Gabor filters with different scales and orientations. Afterwards, we use the edge characteristics associated with the textural and color information as symmetrical weights for voting triangulation. In the end, we construct a polar-based voting histogram based on the accumulation of the symmetry contribution (local texture and color information), in order to find the maximum peaks presenting as candidates of the primary symmetry axes.

The remaining part of the paper is organized as follows. In section 2, we explain Log-Gabor transform and its feature-based application on a gray-scale image. In section 3, textural and color histograms of window-based features are described in details. In section 4, we illustrate the detection of symmetry candidates through triangulation and voting processes. Experimental details and results are presented in section 5. Finally, section 6 contains conclusion and future work.

%%%%%%%%%%%%%%%%%%%%%%%%%%%%%%%%%%%%%%%%%%%%%%%%%%%%%%%%%%%%%%%%%%%%%%%%%%%%%%%%%
\section{Log-Gabor Edge Detection}
%-- Figure: Spatial and Frequency Banks of Log-Gabor -- Done!!
%-- Figure: Output of some input image (Gabor vs Log-Gabor)
Log-Gabor filter consists of logarithmic transformation of a Gabor filter in the Fourier domain, which suppresses the negative effect of the DC component:
\begin{equation}
\hat{G}(\eta,\alpha;s,o) = \hat{G}_s(\eta) \; \hat{G}_o(\alpha)
\end{equation}
\begin{equation}
\hat{G}_s(\eta) = \exp(-\frac{(log(\frac{\eta}{\eta_s}))^2}{2(log(\sigma_\eta))^2}) \: U(\eta)
\end{equation}
\begin{equation}
\hat{G}_o(\alpha) = \exp(-\frac{|{atan(\frac{sin(\alpha - \alpha_o)}{cos(\alpha - \alpha_o)}})|}{2\sigma^{2}_{\alpha}})
\end{equation}
where $(\eta,\alpha)$ are the log-polar coordinates representing radial and angular components over $S$ scales and $O$ orientations, associated with the frequency centers $(\eta_s,\alpha_o)$ and their bandwidths $(\sigma_\eta,\sigma_\alpha)$. $\hat{G}_s(\eta)$ is multiplied by low-pass Butterworth filter $U(\eta)$ of order $15$, and frequency $0.45$, to eliminate any extra frequencies at Fourier corners.

\begin{figure*}[ht!]
	\centering
	\subfloat[Fourier]{\includegraphics[width=0.32\textwidth]{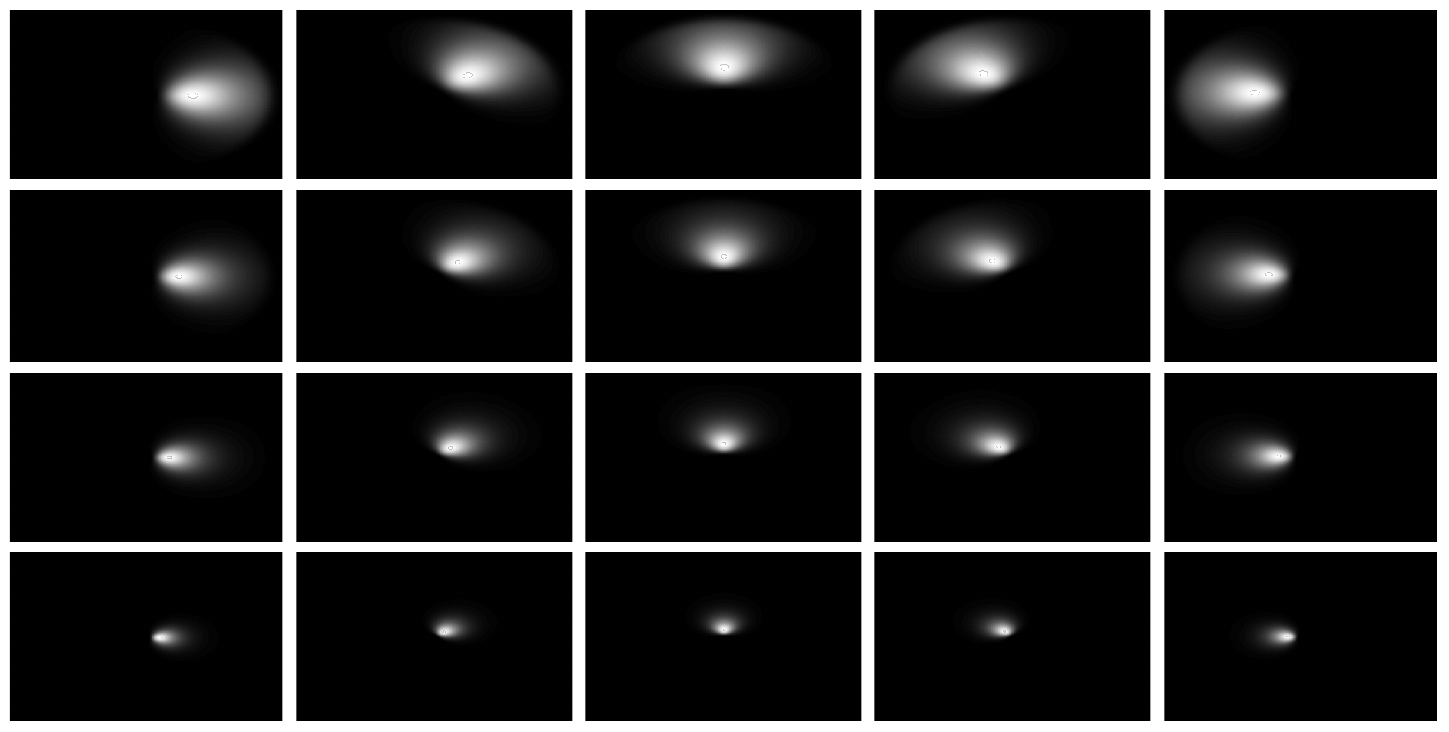}\label{fig:LogGaborF} } 
    \vspace{1.5mm}
	\subfloat[Real]{\includegraphics[width=0.32\textwidth]{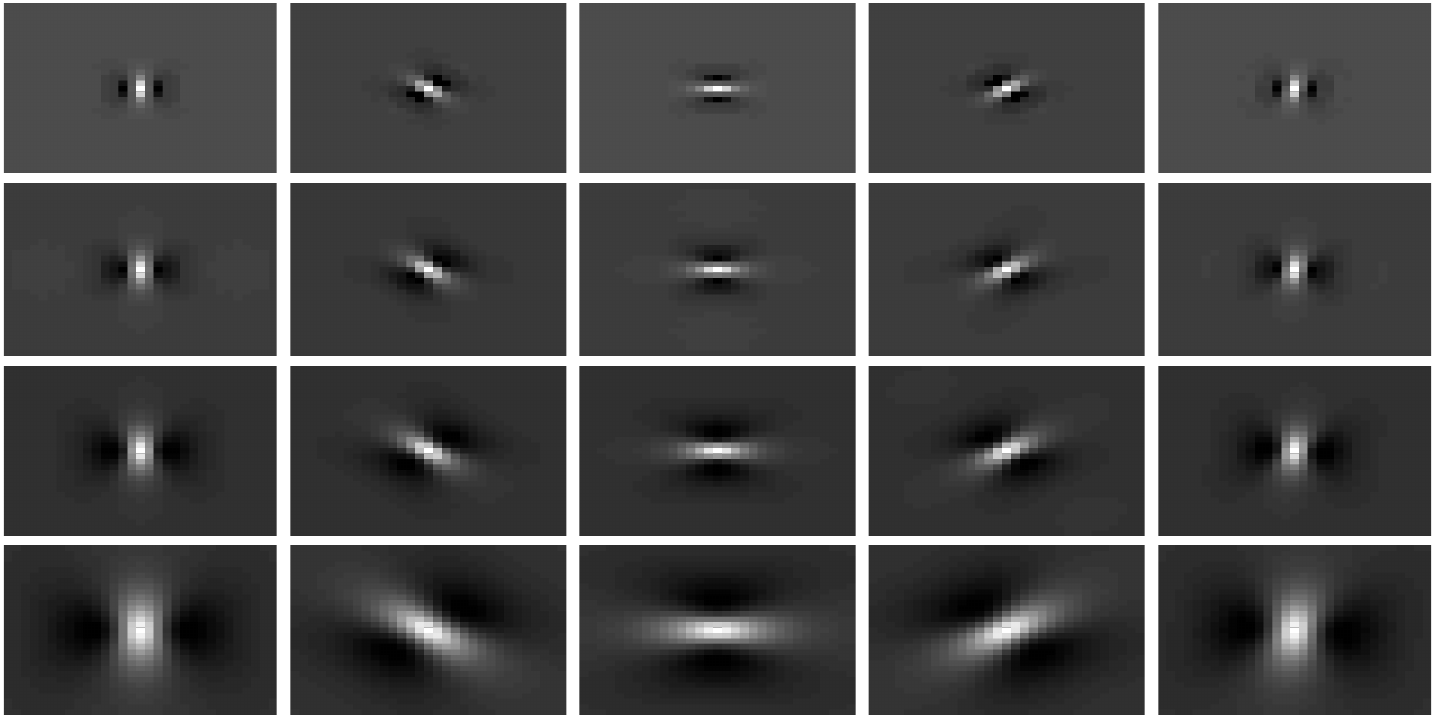}\label{fig:LogGaborR} }
    \vspace{1.5mm}
    \subfloat[Imaginary]{\includegraphics[width=0.32\textwidth]{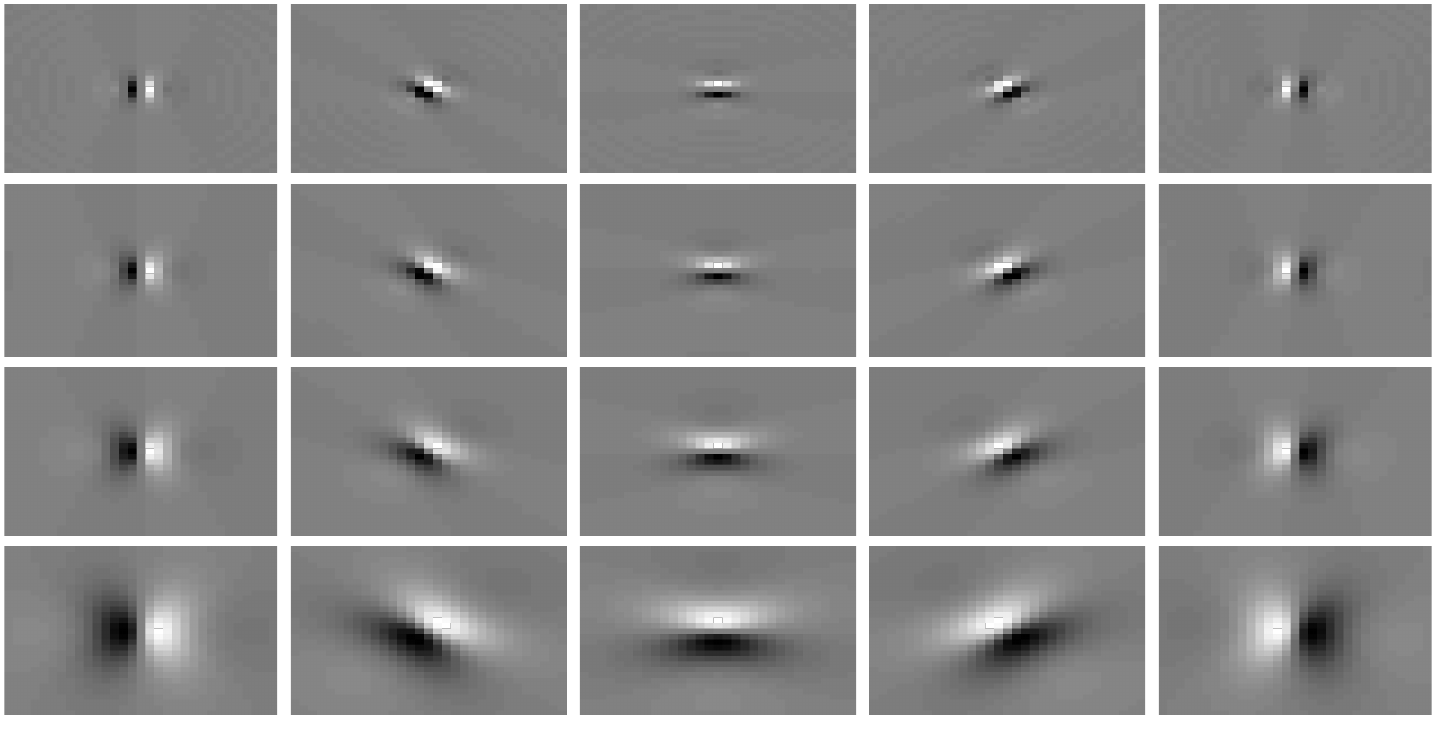}\label{fig:LogGaborI} }
	\caption{Multi-resolution log-Gabor arrangement with $S=4$ scales (in rows) and $O=5$ orientations (in columns). (a) Filters in the Fourier domain. (b,c) Real and imaginary components in the spatial domain.}
	\label{fig:LogGabor}
\end{figure*}

The modulus of complex wavelet coefficients $I_{s,o}(x,y)$ are computed on an image $I$ (width $W$ and height $H$) over multiple scales $s \in \{1,\ldots,S\}$ and orientations $o \in \{\frac{z\pi}{O},z=0,\ldots,O-1\}$ as follows:
\begin{equation}
I \;\; {\underset{GS}\rightarrow} \;\; I_{GS} \;\; {\overset{FT}\rightarrow} \;\; \hat{I}_{GS}
\end{equation}
\begin{equation}
I_{s,o}(x,y) = |{FT}^{-1}(\hat{I}_{GS} \times \hat{G})|
\end{equation}
where $\hat{I}_{GS}$ is the gray-scale version of the image $I$ in frequency domain, and ${FT}(.),\:{FT}^{-1}(.)\;$ are the Fourier transform and its inverse. Figure~\ref{fig:LogGaborF} shows Log-Gabor wavelet filter bank with $4$ scales and $5$ orientations where schematic contours cover the frequency space in the Fourier domain. Consequently, the elongation scheme of Log-Gabor wavelets appears in figures~\ref{fig:LogGaborR},~\ref{fig:LogGaborI} presenting the real and imaginary components in the spatial domain.

\begin{figure*}[]
	\centering
	\subfloat[Input]{\includegraphics[width=0.31\textwidth]{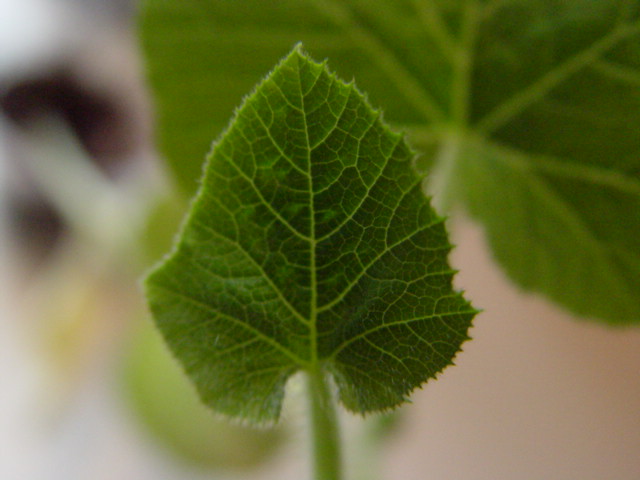}\label{fig:lg_inp} } 
    \vspace{1.5mm}
	\subfloat[Amplitude]{\includegraphics[width=0.32\textwidth]{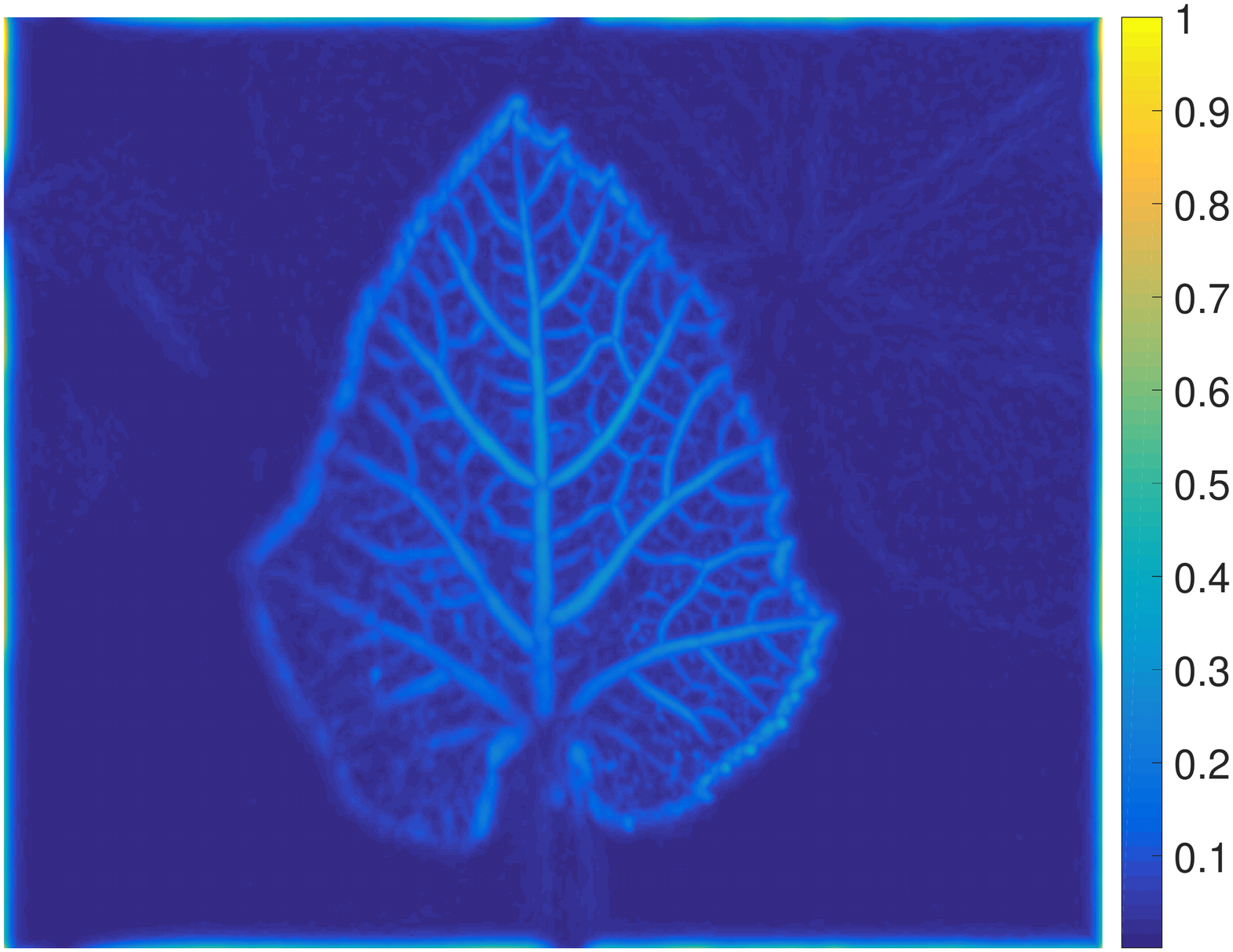}\label{fig:lg_amp} }
    \vspace{1.5mm}
    \subfloat[Orientation]{\includegraphics[width=0.32\textwidth]{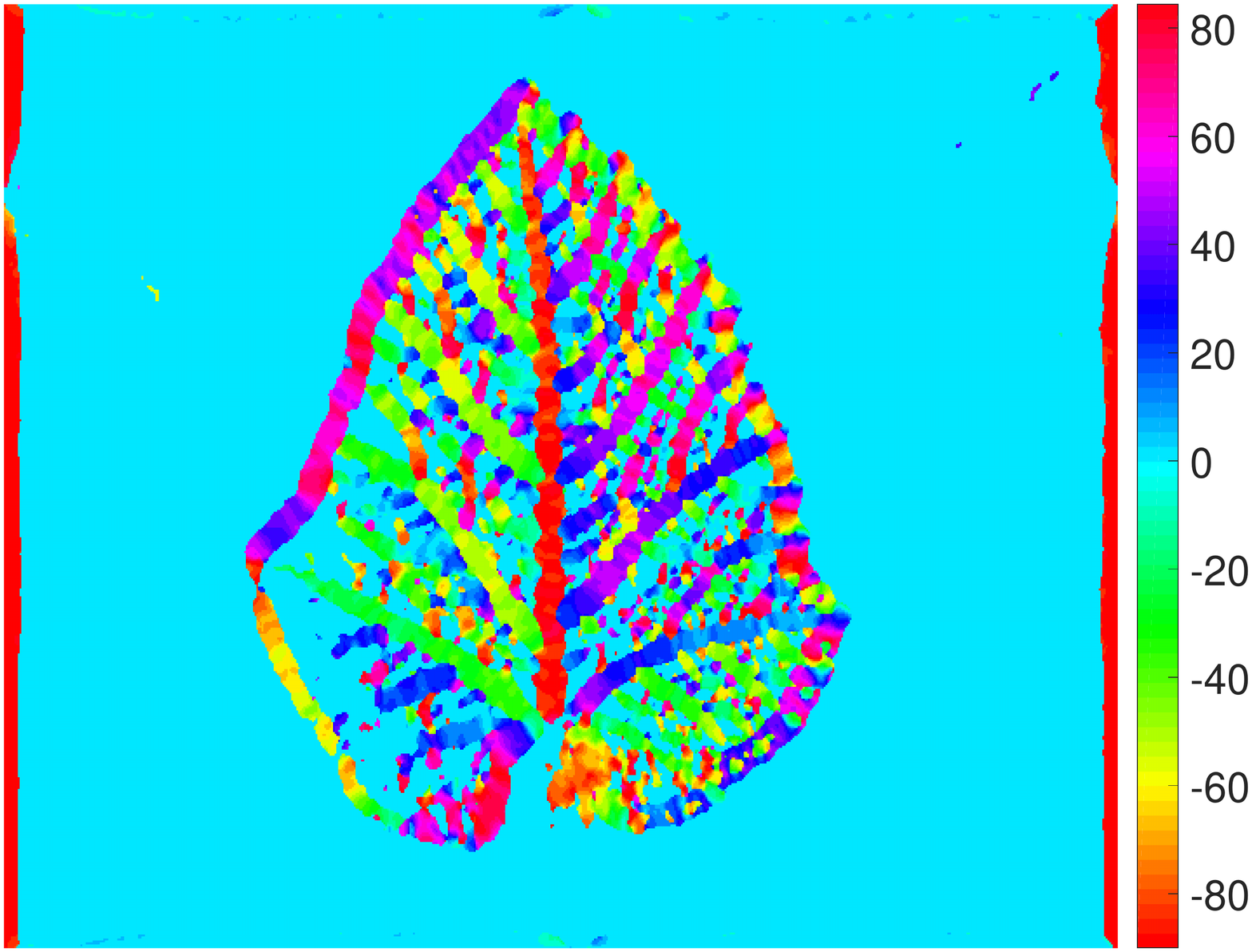}\label{fig:lg_ang} }
	\caption{Log-Gabor response: (a) Input image. (b) Amplitude map $J \in [0,1]$. (c) Orientation map $\phi \in [-90^{\circ},90^{\circ}]$.}
	\label{fig:LogGabor_Res}
\end{figure*}

Figure~\ref{fig:LogGabor_Res} presents an example of Log-Gabor response $I_{s,o}$ on some natural image with a blurring background. Amplitude map $J(x,y) = \max\limits_{s,o} I_{s,o}(x,y)$ highlights the edge details of the foreground object in a sharp way, accompanied by precise angular values in the corresponding orientation map $\phi(x,y)$. Upon a spatial sampling of the input image $I$ using non-interleaved cells along a regular grid (stride and cell size are proportional to the maximum image dimension $\max(W,H)$). A feature point $p^i$ is extracted within each non-homogeneous cell $D_i$ using the wavelet response of Log-Gabor filter $I_{s,o}(D_i)$, associated with its maximum wavelet response $J_i = J(p^i)$ along side with the corresponding orientation $\phi_i$ and color in $HSV$ color space $\psi_i$. 

%%%%%%%%%%%%%%%%%%%%%%%%%%%%%%%%%%%%%%%%%%%%%%%%%%%%%%%%%%%%%%%%%%%%%%%%%%%%%%%%%
\section{Textural and Color Histograms}
The textural and color information around an edge segment are prominent similarity characteristics for natural images, describing the symmetrical behavior of local edge orientation, and the balanced distribution of luminance and chrominance components. Hence, we introduce two histograms: Firstly,  neighboring textural histogram $h^i$ of size $N$:
\begin{equation}
h^i(n) = \sum_{r \in D_i} J_r \; \mathbbm{1}_{\Phi_n}(\phi_r), 
\end{equation}
\vspace{-3mm}
\begin{align*}
\Phi_n = [\frac{n\pi}{N}, \frac{(n+1)\pi}{N}[, \; n = 0,\ldots, N-1
\end{align*}
where $\mathbbm{1}$ is the indicator function. $h^i$ is $l1$ normalized and circular shifted with respect to the orientation of the maximum magnitude $\phi_i$ among the neighborhood cell $D_i$. Secondly, the local $HSV$ histogram $g^i$ of size $C$ with sub-sampling rate ($C^{hu}:C^{sa}:C^{va}$)
\begin{equation}
    g^i(c) = \sum_{r \in D^*_i} \mathbbm{1}_{\Psi_c}(\psi_r),
\end{equation}
\vspace{-5mm}
\begin{align*}
c &= (c^{hu},c^{sa},c^{va}), \\
c^{hu} &\in \{0, \ldots, C^{hu}-1\}, \\
c^{sa} &\in \{0, \ldots, C^{sa}-1\}, \\
c^{va} &\in \{0, \ldots, C^{va}-1\},
\end{align*}
\vspace{-6.5mm}
\begin{gather*}
\scalebox{1.04}{$\Psi_c = [\frac{2c^{hu}\pi}{C^{hu}},\frac{2(c^{hu}+1)\pi}{C^{hu}}[ \: \times \: [ \frac{c^{sa}}
{C^{sa}}, \frac{c^{sa}+1}{C^{sa}} [ \: \times \: [ \frac{c^{va}}{C^{va}}, \frac{c^{va}+1}{C^{va}} [ $}
\end{gather*}
where $D^*_i$ is the neighborhood window around feature point $p^i$ , $\psi_c$ is a sub-sampled set of $HSV$ space, in terms of three components: hue ($hu$), saturation ($sa$) and value ($va$). $l1$ normalization is applied to $g^i(.)$ for bin-wise histogram comparison.
%%%%%%%%%%%%%%%%%%%%%%%%%%%%%%%%%%%%%%%%%%%%%%%%%%%%%%%%%%%%%%%%%%%%%%%%%%%%%%%%%
\section{Symmetry Triangulation and Voting}
%-- Figure: triangulation process
%-- Figure: step-by-step example
A set of feature pairs $(p^{i},p^{j})$ of size $\frac{P(P-1)}{2}$ are elected such that $i \neq j$, and $P$ is the number of feature points. Then, we compute the symmetry candidate axis based a triangulation process with respect to the image origin. This candidate axis is parametrized by angle $\theta_{i,j}$ (orientation of the norm), and displacement $\rho_{i,j}$ (norm distance to the image origin) and has a symmetry weight $\omega_{i,j}$  defined as follows:

\begin{equation}
\omega_{i,j} = \omega(p^{i},p^{j}) =  m(i,j) \; t(i,j) \; q(i,j)
\label{eq:weights}
\end{equation}
\begin{align}
    m(i,j) &= |\tau^i R(T_{ij}^\perp) \tau^j| \\
    t(i,j) &= \sum_{n=1}^{N} min(h^i(n),\tilde{h}^j(n)) \\
    q(i,j) &= \sum_{c=1}^{C} min(g^i(c),g^j(c))
\end{align}
where $\tau^i$ is a direction vector associated with angle $\phi_i$, $R(T_{ij}^\perp)$ is the reflection matrix with respect to the perpendicular of the line connecting $(p^{i},p^{j})$ \cite{Cicconet2014,Elawady2016}, and $\tilde{h}^j$ is the reverse version of $h^j$ histogram. $l1$ normalization is applied to symmetry weights $\omega$.

\begin{figure}[]
	\centering
	\subfloat[Input with GT]{\includegraphics[width=0.46\columnwidth ]{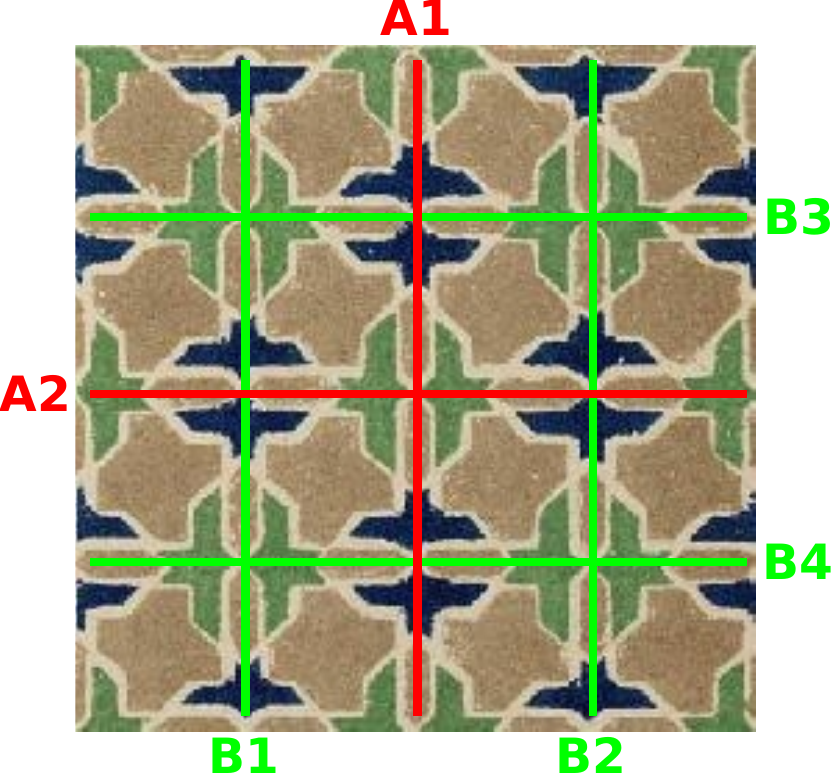}\label{fig:vote_input} }
    \subfloat[Voting representation]{\includegraphics[width=0.53\columnwidth ]{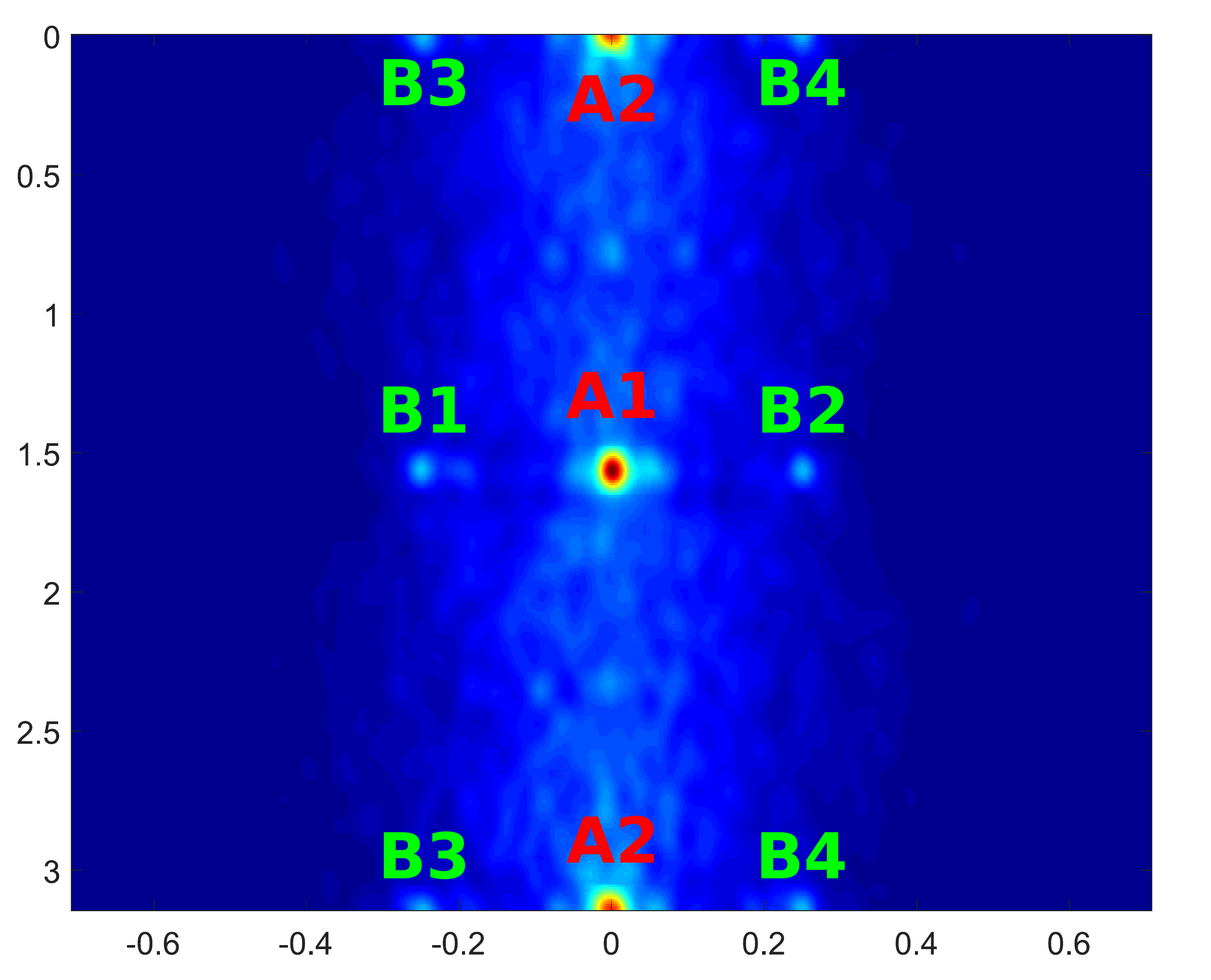}\label{fig:vote_all} }
	\caption{Symmetry voting process: (a) Input image with major (in red) and minor (in green) symmetry axes. (b) Smoothed output of the symmetry histogram $H$ with $\rho$ displacement x-axis and $\theta$ angle y-axis, highlighting the corresponding axes.}
	\label{fig:vote}
\end{figure}

A symmetry histogram $H(\rho,\theta)$ is defined as the sum of the symmetry weights of all pairs of feature points such as:
\begin{equation}
H(\rho,\theta)  = \sum_{\substack{p^i,p^j\\i\neq{j}}} \omega_{i,j} \delta_{\rho - \rho_{i,j}}  \delta_{\theta - \theta_{i,j}} 
\end{equation}
where $\delta$ is the Kronecker delta.

the voting histogram $H(\rho,\theta)$ is smoothed using a Gaussian kernel to output a proper density representation (as shown in figure~\ref{fig:vote}), in which the major symmetry peaks are selected by reaching-out clear extreme spots using well-known non-maximal suppression technique \cite{canny1986computational}. The spatial boundaries of each symmetry axis is computed through the convex hull of the associated voting pairs.

%%%%%%%%%%%%%%%%%%%%%%%%%%%%%%%%%%%%%%%%%%%%%%%%%%%%%%%%%%%%%%%%%%%%%%%%%%%%%%%%%
\section{Results and Discussions}
This section describes the details of public symmetry datasets, evaluation metrics, and experimental settings for performance comparison of the proposed work. Performance analysis against the state-of-the-art algorithms is also provided in this section.
%%%%%%%%%%%%%%%%%%%%%%%%%%%%%%%%%%%%%%%%%%%%%%%%%%%%%%%%%%%%%%%%%%%%%%%%%%%%%%%%%
\subsection{Datasets description}
Seven public datasets of reflection symmetry detection are used from four different databases: (1) PSU datasets (single,multiple): Liu's vision group proposed symmetry groundtruth  for Flickr images (\# images = \# symmetries = 157 for single case, \# images = 142 and \# symmetries = 479 for multiple case) in ECCV2010\footnote[1]{http://vision.cse.psu.edu/research/symmetryCompetition/index.shtml}, CVPR2011\footnote[2]{http://vision.cse.psu.edu/research/symmComp/index.shtml} and CVPR2013\footnote[3]{http://vision.cse.psu.edu/research/symComp13/content.html}. (2) AVA dataset (single): Elawady et. al \cite{Elawady2016} provided axis groundtruth\footnote[4]{http://github.com/mawady/AvaSym} for some professional photographs (\# images = \# symmetries = 253 for single case) from large-scale aesthetic-based dataset called AVA \cite{Murray2012}.  (3) NY datasets (single,multiple): Cicconet et al. \cite{Cicconet2016} introduced a symmetry database (\# images = \# symmetries = 176 for single case, \# images = 63 and \# symmetries = 188 for multiple case) in 2016\footnote[5]{http://symmetry.cs.nyu.edu/}, providing more stable groundtruth. (4) ICCV2017 training datasets (single,multiple): Seungkyu Lee delivered a challenge database associated with reflection groundtruth\footnote[6]{https://sites.google.com/view/symcomp17/challenges/2d-symmetry} (\# images = \# symmetries = 100 for single case, \# images = 100 and \# symmetries = 384 for multiple case).
%%%%%%%%%%%%%%%%%%%%%%%%%%%%%%%%%%%%%%%%%%%%%%%%%%%%%%%%%%%%%%%%%%%%%%%%%%%%%%%%%
\subsection{Evaluation metrics}
Assuming a symmetry line defined by two endpoints $(a=[a_x,x_y]^T,\;b=[b_x,b_y]^T)$, quantitative comparisons are fundamentally performed by considering a detected symmetry candidate $SC = [a^{SC},b^{SC}]^T$ as a true positive (TP) respect to the corresponding groundtruth $GT = [a^{GT},b^{GT}]^T$ if satisfying the following two conditions:  
\begin{equation}
   	T({atan}(\frac{{abs(\begin{vmatrix} v_x^{SC} & v_x^{GT} \\ v_y^{SC} & v_y^{GT} \end{vmatrix})}}{<v^{SC}, v^{GT}>})) < \gamma \: ,
\end{equation}
\begin{equation}
    \sqrt{(t_x^{SC}-t_x^{GT})^2+(t_y^{SC}-t_y^{GT})^2} < \zeta \: ,
\end{equation}
\begin{equation*}
   	v^{SC} = (a^{SC}-b^{SC}), \: v^{GT} = (a^{GT}-b^{GT}) \: ,
\end{equation*}
\begin{equation*}
   	t^{SC} = \frac{(a^{SC}+b^{SC})}{2}, \: t^{GT} = \frac{(a^{GT}+b^{GT})}{2},
\end{equation*}
\begin{equation*}
   	T(\Gamma) =
\begin{cases}
    \pi-\Gamma,& \text{if } \Gamma > \frac{\pi}{2}\\
    \Gamma,              & \text{otherwise}
\end{cases}
\end{equation*}

The conditions represent angular and distance constraints between detection and groundtruth axes. These constraints are upper-bounded by the corresponding thresholds $\gamma$ and $\zeta$, which are defined in table~\ref{tab:Eval}. Furthermore, the precision $PR$ and recall $RC$ rates are defined by selecting the symmetry peaks according to the candidates' amplitude normalized by the highest detection score, to show the performance curve for each algorithm:
\begin{equation}
PR = \frac{TP}{TP+FP}, \: RC = \frac{TP}{TP+FN}
\end{equation}
where $FP$ is a false positive (non-matched detection), and $FN$ is a false negative (non-matched groundtruth). In addition, we used the maximum $F_1$ score identifying a unique comparison measure, to show the overall accuracy of each symmetry algorithm:

\begin{equation}
{max}\{F_1\} = {max}\{2\frac{PR \times RC}{PR + RC}\}
\end{equation}

\begin{table}[h]
\centering
\caption{Threshold values of evaluation metrics across different reflection symmetry competitions.}
\label{tab:Eval}
\begin{tabular}{@{}c|c|c@{}}
\hline
\hline
{ \small Competitions} & { \small $\gamma$} & { \small $\zeta$} \\
\hline
\hline
{ \small CVPR2011 \cite{Rauschert2011}} & $10^{\circ}$  & $20\% \times {len}(GT)$  \\
{ \small CVPR2013 \cite{Liu2013}}  & $10^{\circ}$ & $20\% \times {min}\{{len}(MT),{len}(GT)\}$ \\
{ \small ICCV2017} & $3^{\circ}$ & $2.5\% \times {min}\{W,H\}$  \\
\hline
\hline
\end{tabular}
\end{table}
%%%%%%%%%%%%%%%%%%%%%%%%%%%%%%%%%%%%%%%%%%%%%%%%%%%%%%%%%%%%%%%%%%%%%%%%%%%%%%%%%
\subsection{Experimental settings}
We compare the proposed methods (\textit{Lg}: without color information and \textit{LgC}: with color information) against three state-of-the-art approaches: Loy and Eklundh (\textit{Loy2006}) \cite{Loy2006}, Cicconet et al. (\textit{Cic2014}) \cite{Cicconet2014}, and Elawady et al. (\textit{Ela2016}) \cite{Elawady2016}). Their open-source codes are used with default parameter values for performance comparison. In Log-Gabor edge detection, we set the number of scales $S$ and number of orientations $O$ to $12$ and $32$. We also set the radial bandwidth $\sigma_\eta$ to $0.55$ and the angular bandwidth $\sigma_\alpha$ to $0.2$. In textural and color histogram calculations, we define the number of bins for textural $N$ and color $C$ to $32$ and $32$ (sampling rate $C^{hu}:C^{sa}:C^{va} = 8:2:2$) respectively. In case of gray-scale images, contrast values are used instead of color information in $HSV$ color space. In symmetry voting, we declare the 2D histogram space of $\rho=\sqrt{W^2+H^2}$ displacement bins and $\theta=360$ orientation bins for extrema selection.
%%%%%%%%%%%%%%%%%%%%%%%%%%%%%%%%%%%%%%%%%%%%%%%%%%%%%%%%%%%%%%%%%%%%%%%%%%%%%%%%%
\subsection{Performance analysis}
%-- Figure: Symmetry examples (single/multiple)
\begin{figure*}[h]
	\centering
	\subfloat[PSU]{\includegraphics[width=0.32\textwidth]{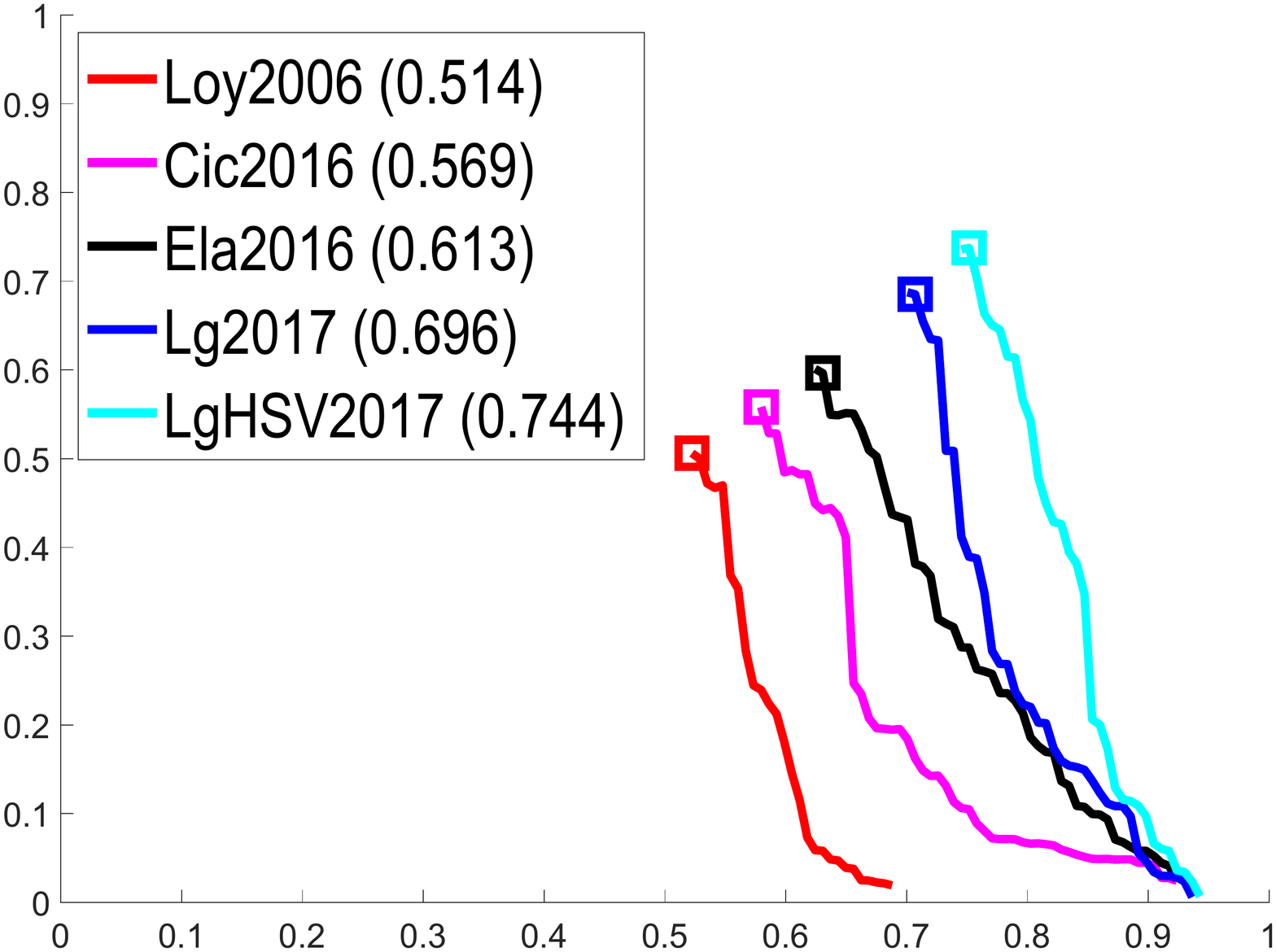}\label{fig:prCurve_PSU} } 
    \subfloat[AVA]{\includegraphics[width=0.32\textwidth]{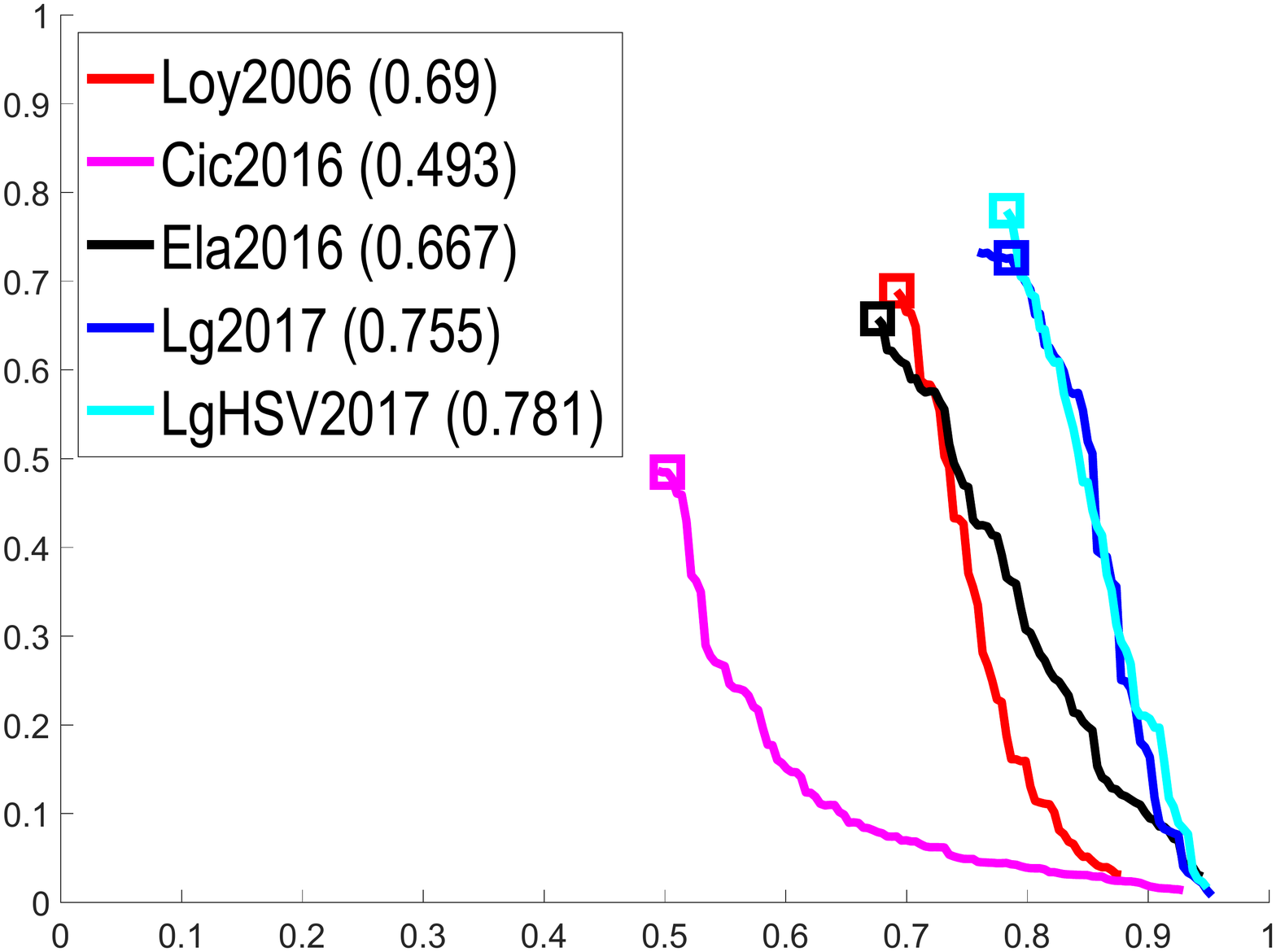}\label{fig:prCurve_AVA} }
	\subfloat[NY]{\includegraphics[width=0.32\textwidth]{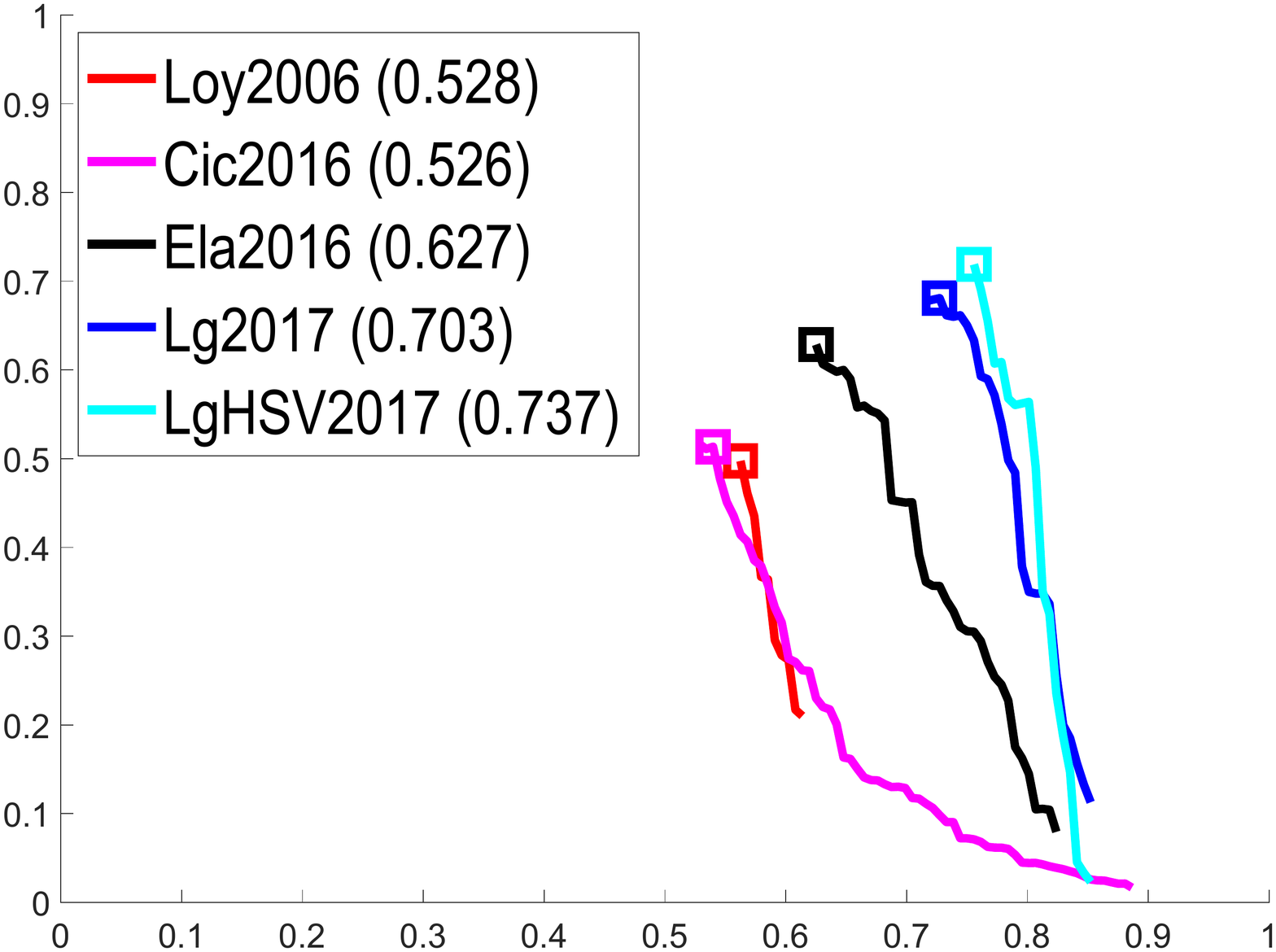}\label{fig:prCurve_NY} } 
    \\
   \subfloat[ICCV'17]{\includegraphics[width=0.32\textwidth]{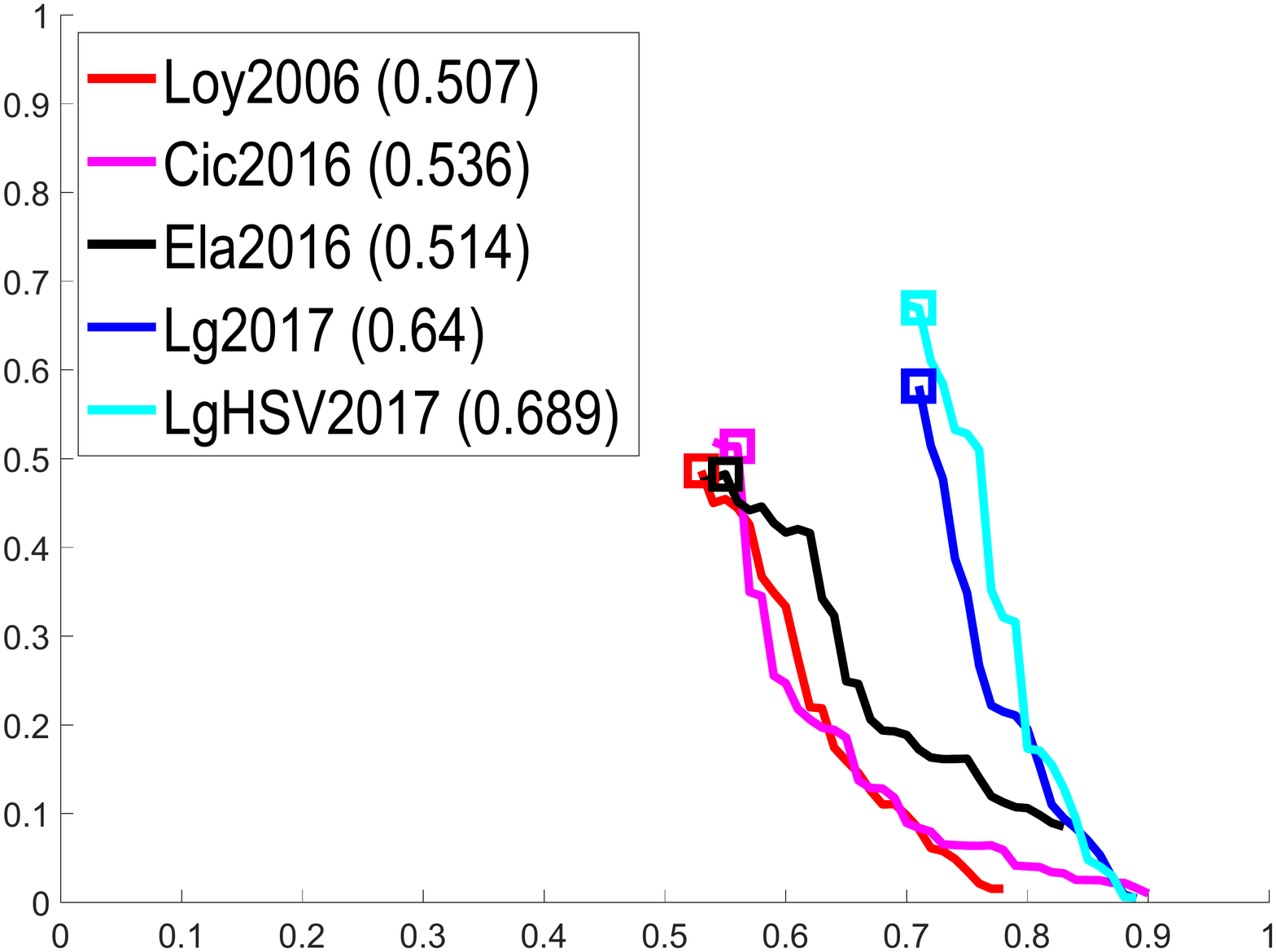}\label{fig:prCurve_ICCV17} }
    \subfloat[PSUm]{\includegraphics[width=0.32\textwidth]{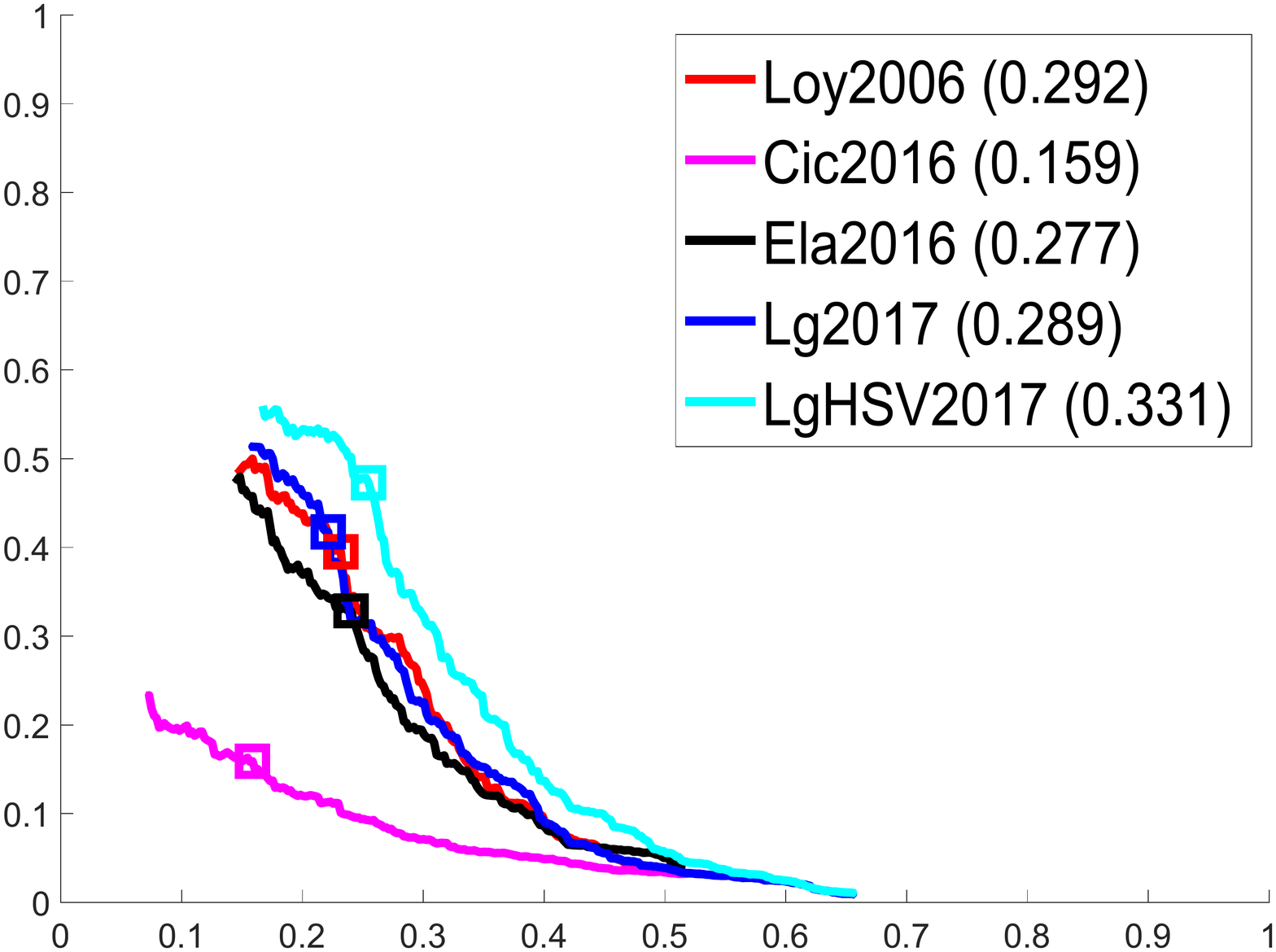}\label{fig:prCurve_PSUm} } 
	\subfloat[NYm]{\includegraphics[width=0.32\textwidth]{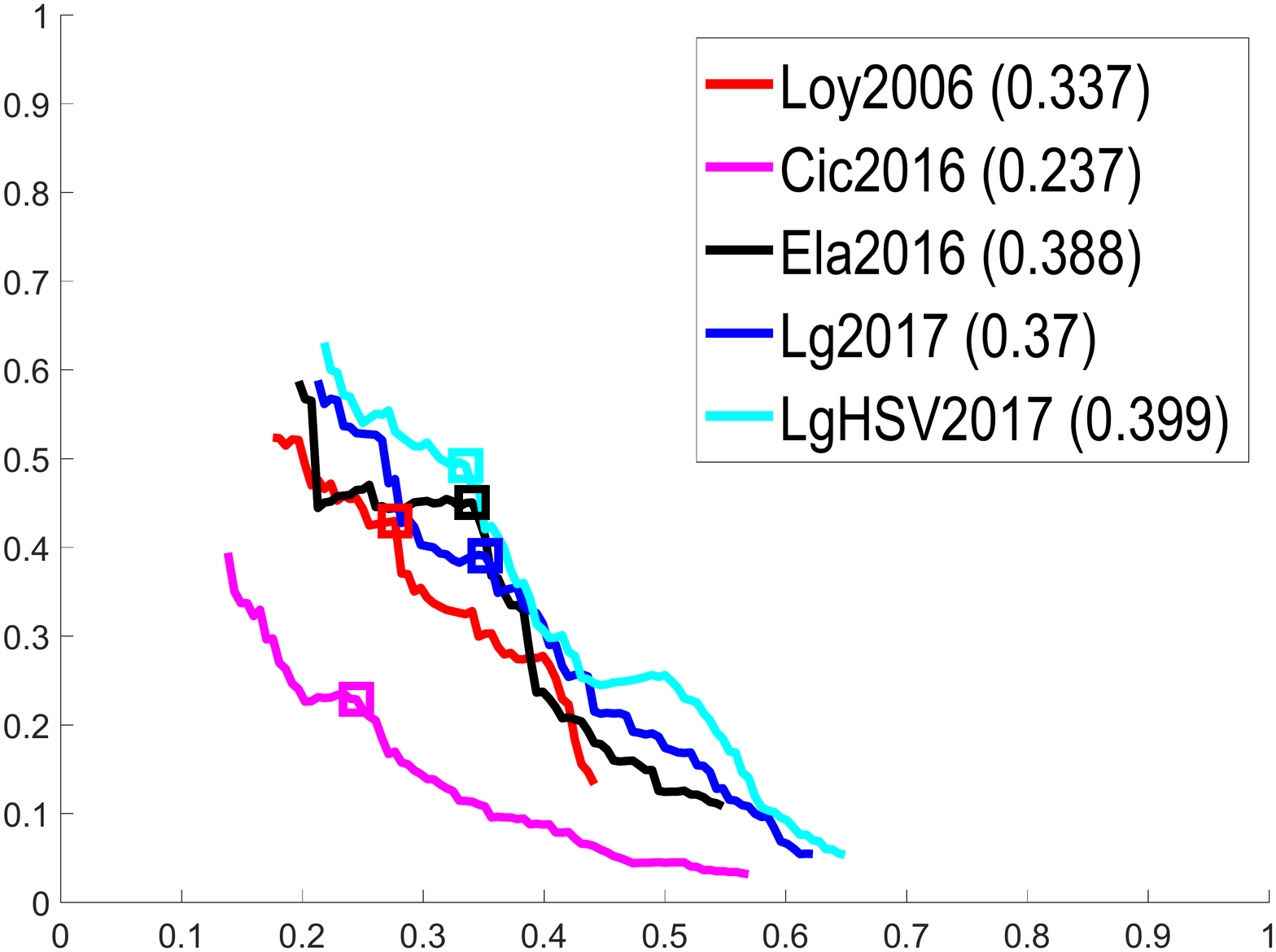}\label{fig:prCurve_NYm} }
    \\
    	\subfloat[ICCV'17m]{\includegraphics[width=0.32\textwidth]{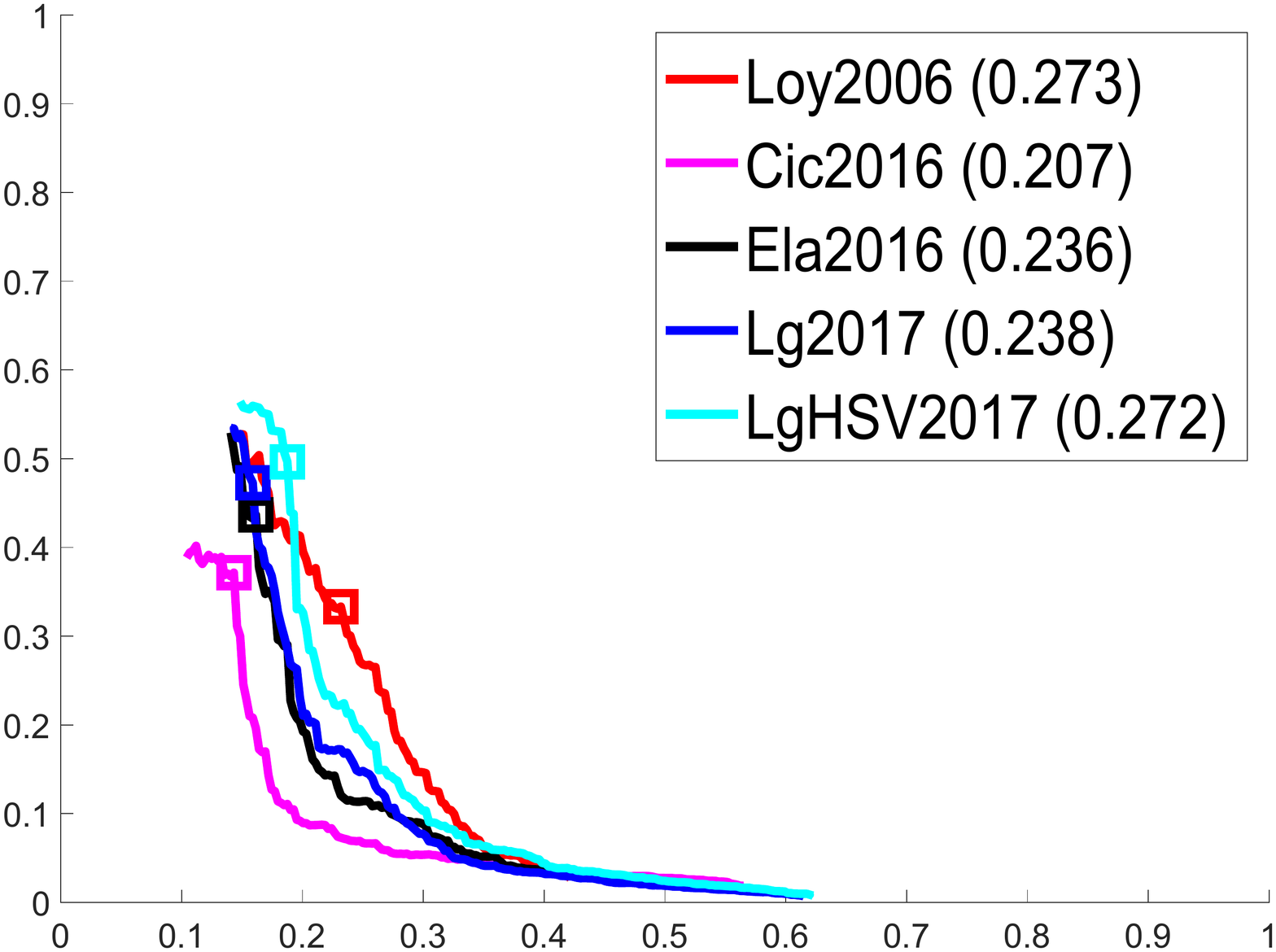}\label{fig:prCurve_ICCV2017m} } 
	\caption{Using evaluation metrics \textbf{CVPR2013} \cite{Liu2013}, Precision-Recall curves on: (1) four single-case symmetry (a,b,c,d) datasets, and (2) three multiple-case symmetry (e,f,g) datasets to show the overall superior performance of our proposed methods ( \textit{Lg2017} and \textit{LgHSV2017}) against the three prior algorithms (\textit{Loy2006} \cite{Loy2006}, \textit{Cic2014} \cite{Cicconet2014}, and \textit{Ela2016} \cite{Elawady2016}). The maximum F1 scores are qualitatively presented as square symbols along the curves, and quantitatively indicated between parentheses inside the top-right legends. Best seen on screen (zoom-in for details).}
	\label{fig:prCurve}
\end{figure*}

\begin{table}[h]
\centering
\caption{Using evaluation metrics \textbf{CVPR2013} \cite{Liu2013}, comparison of the true positive rates based on top detection for the proposed methods against the state-of-art algorithms. Symmetry datasets are presented as: single-case (first 4 rows) and multiple-case (last 3 rows), highlighted between (parenthesis) the number of images for each dataset. Top 1st and 2nd values are in \textbf{Bold} and \underline{underlined} respectively.}
\label{tab:resTP}
%\vspace{3mm}
\begin{tabular}{@{}c|c|c|c|c|c@{}}
\hline
\hline
{ \small Datasets} & { \small \textit{Loy}\cite{Loy2006}} & { \small \textit{Cic}\cite{Cicconet2014}} & { \small \textit{Ela}\cite{Elawady2016}} & { \small \textit{Lg}} & { \small \textit{LgC}} \\
\hline
\hline
{ \footnotesize PSU(157)} & 81 & 90 & 97 & \underline{109} & \textbf{116} \\
{ \footnotesize AVA(253)} & 174 & 124 & 170 & \underline{191} & \textbf{197} \\
{ \footnotesize NY(176)}  & 98 & 92 & 109 & \underline{125} & \textbf{132} \\
{ \footnotesize ICCV17(100)} & 52 & 53 & 52 & \textbf{70} & \underline{69} \\
\hline
{ \footnotesize PSUm(142)}  & 69 & 68 & 67 & \underline{74} & \textbf{79} \\
{ \footnotesize NYm(63)}  & 32 & 36 & 36 & \underline{39} & \textbf{40} \\
{ \footnotesize ICCV17m(100)} & \underline{54} & 39 & 53 & 53 & \textbf{56} \\
\hline
\hline
\end{tabular}
\end{table}

\begin{table}[h]
\centering
\caption{Using evaluation metrics \textbf{ICCV17}, comparison of the true positive rates based on top detection in single-case and all detections in multiple-case for the proposed methods against the state-of-art algorithms. Symmetry datasets are presented as: single-case (first 4 rows) and multiple-case (last 3 rows), highlighted between (parenthesis) the number of groundtruth for each dataset. Top 1st and 2nd values are in \textbf{Bold} and \underline{underlined} respectively.}
\label{tab:resTPR17}
%\vspace{3mm}
\begin{tabular}{@{}c|c|c|c|c|c@{}}
\hline
\hline
{ \small Datasets} & { \small \textit{Loy}\cite{Loy2006}} & { \small \textit{Cic}\cite{Cicconet2014}} & { \small \textit{Ela}\cite{Elawady2016}} & { \small \textit{Lg}} & { \small \textit{LgC}} \\
\hline
\hline
{ \footnotesize PSU(157)} & 41 & 27 & 46 & \textbf{52} & \underline{51} \\
{ \footnotesize AVA(253)} & 63 & 36 & \textbf{95} & 86 & \underline{87} \\
{ \footnotesize NY(176)}  & 34 & 35 & 43 & \underline{50} & \textbf{53} \\ 
{ \footnotesize ICCV17(100)} & 17 & 9 & \underline{19} & \textbf{22} & \textbf{22} \\
\hline
{ \footnotesize PSUm(479)} & 130 & 65 & 129 & \underline{154} & \textbf{157} \\
{ \footnotesize NYm(188)} & 48 & 45 & \textbf{69} & 57 & \underline{60} \\
{ \footnotesize ICCV17m(384)} & \underline{95} & 38 & 86 & \textbf{111} & \textbf{111} \\
\hline
\hline
\end{tabular}
\end{table}

\begin{figure*}
	\centering
    \subfloat[16550 - GT]{\includegraphics[width=0.32\columnwidth]{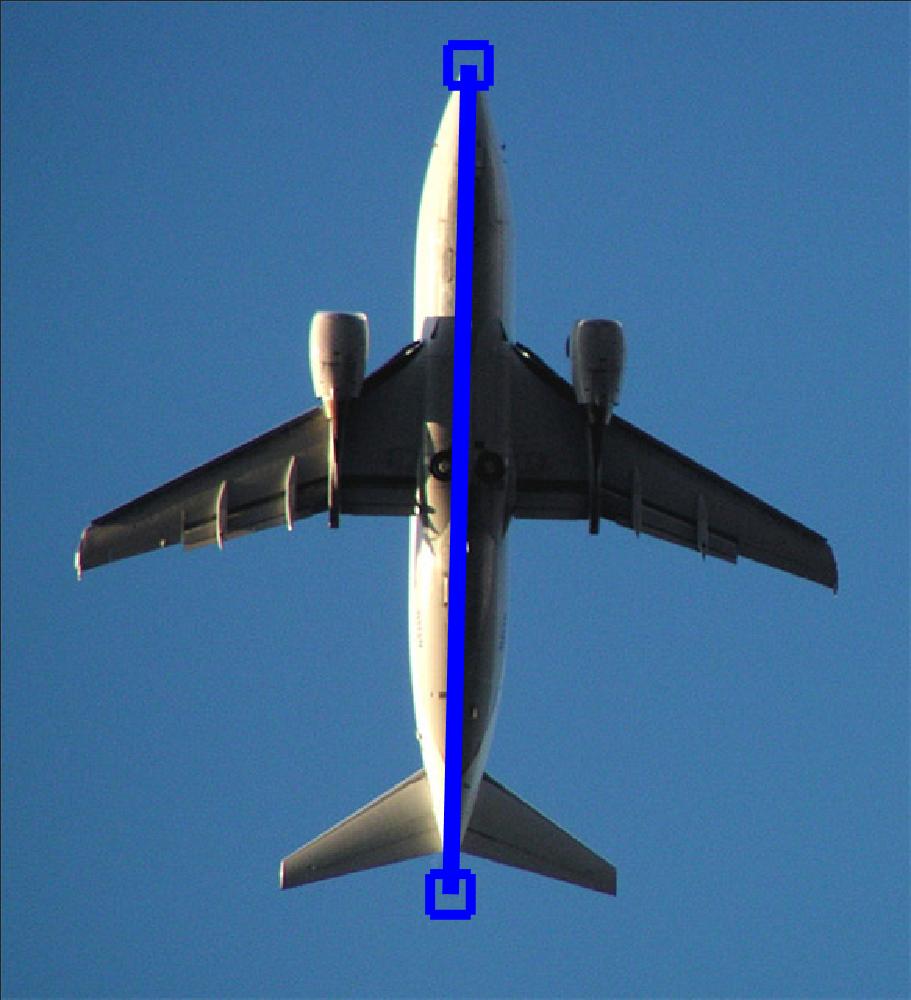}\label{fig:AVA1_GT} } 
    \hspace{1mm}
    \subfloat[832486 - GT]{\includegraphics[width=0.4\columnwidth]{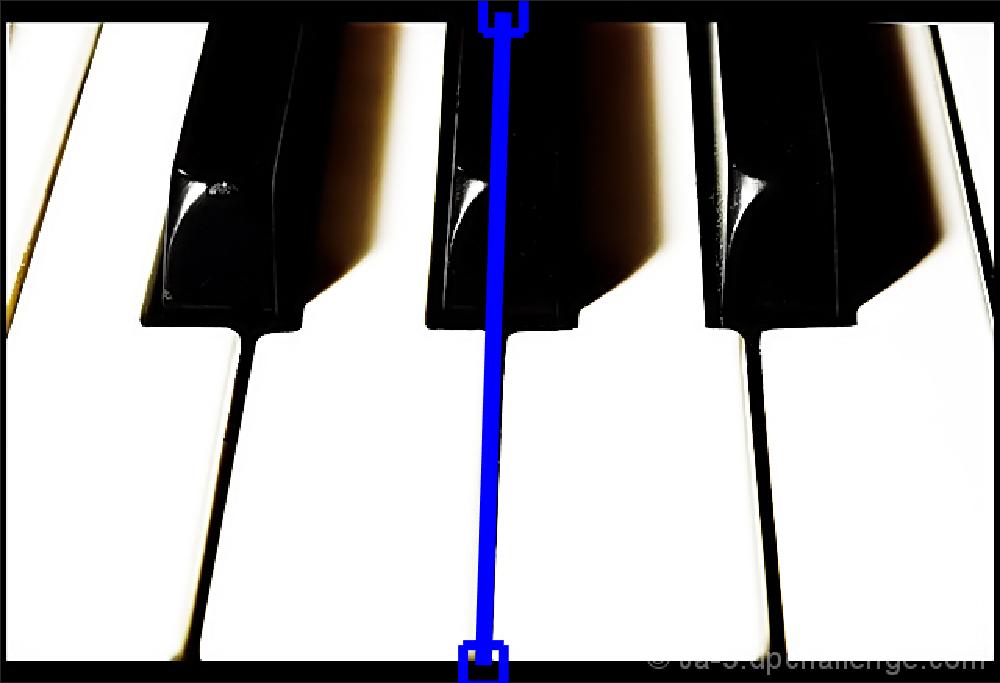}\label{fig:AVA2_GT} } 
    \hspace{1mm}
    \subfloat[ref\_rm\_13 - GT]{\includegraphics[width=0.39\columnwidth]{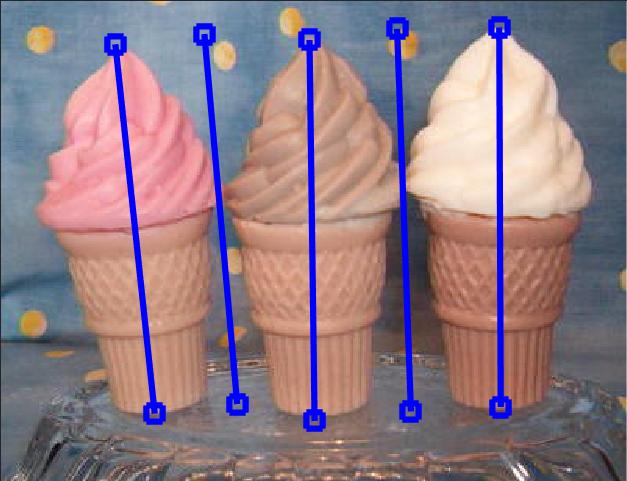}\label{fig:PSUm1_GT} } 
    \hspace{1mm}
    \subfloat[ref\_rm\_65 - GT]{\includegraphics[width=0.4\columnwidth]{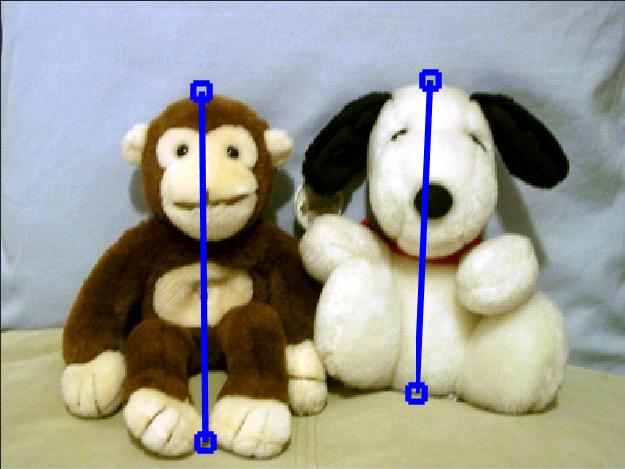}\label{fig:PSUm2_GT} } 
    \hspace{1mm}
    \subfloat[I034 - GT]{\includegraphics[width=0.4\columnwidth]{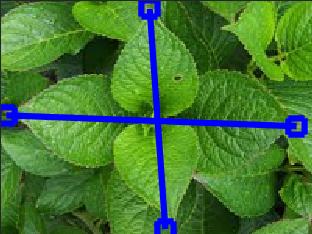}\label{fig:NYm1_GT} } 
    	\\
	\subfloat[16550 - LgC]{\includegraphics[width=0.32\columnwidth]{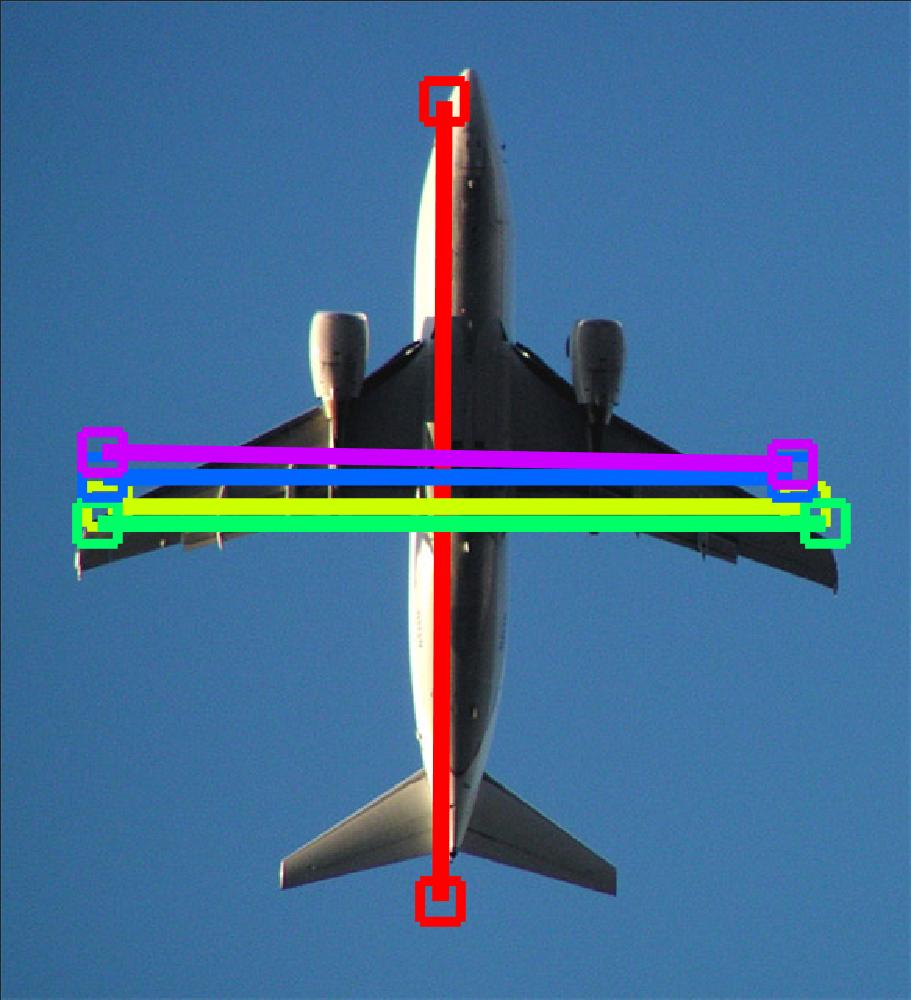}\label{fig:AVA1_OurHSV} }
    \hspace{1mm}
    \subfloat[832486 - LgC]{\includegraphics[width=0.4\columnwidth]{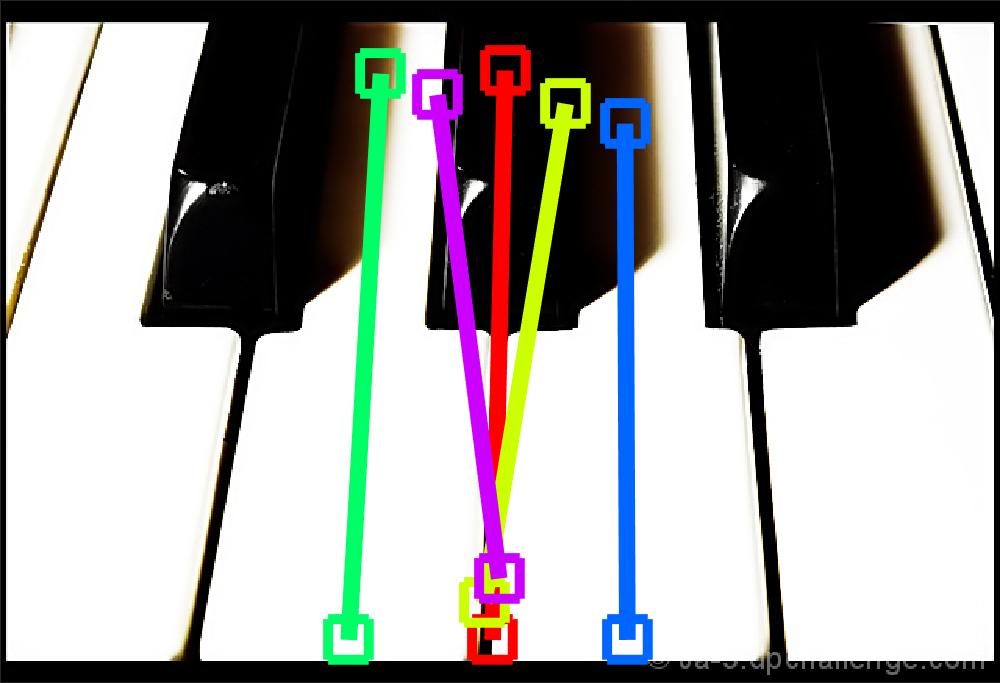}\label{fig:AVA2_OurHSV} }
    \hspace{1mm}
    \subfloat[ref\_rm\_13 - LgC]{\includegraphics[width=0.39\columnwidth]{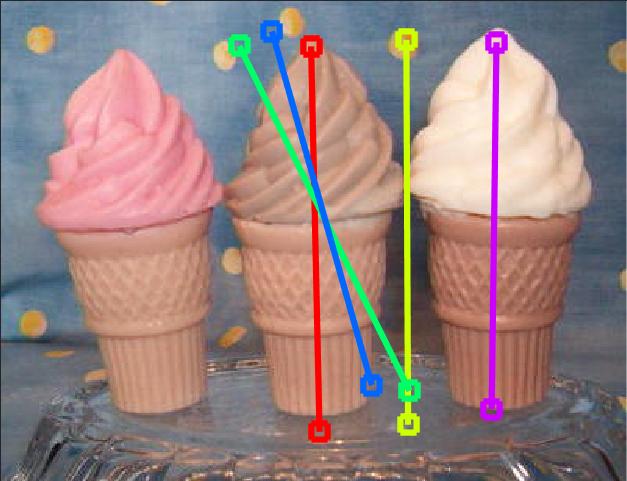}\label{fig:PSUm1_OurHSV} } 
    \hspace{1mm}
    \subfloat[ref\_rm\_65 - LgC]{\includegraphics[width=0.4\columnwidth]{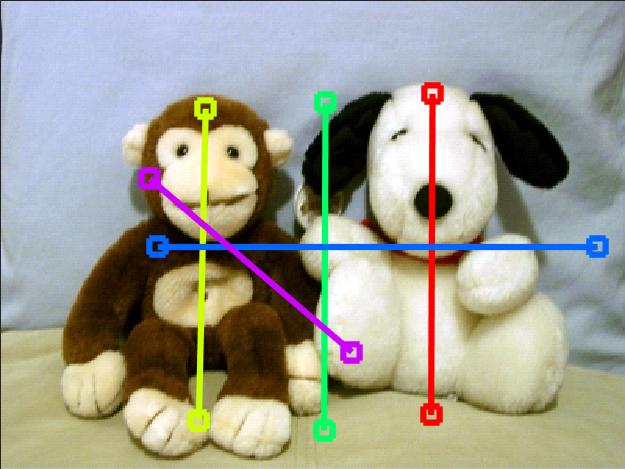}\label{fig:PSUm2_OurHSV} } 
    \hspace{1mm}
    \subfloat[I034 - LgC]{\includegraphics[width=0.4\columnwidth]{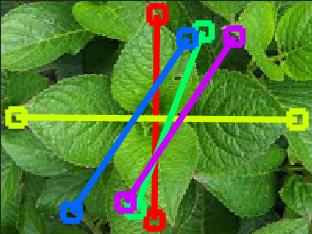}\label{fig:NYm1_OurHSV} }
    \\
    \subfloat[16550 - Lg]{\includegraphics[width=0.32\columnwidth]{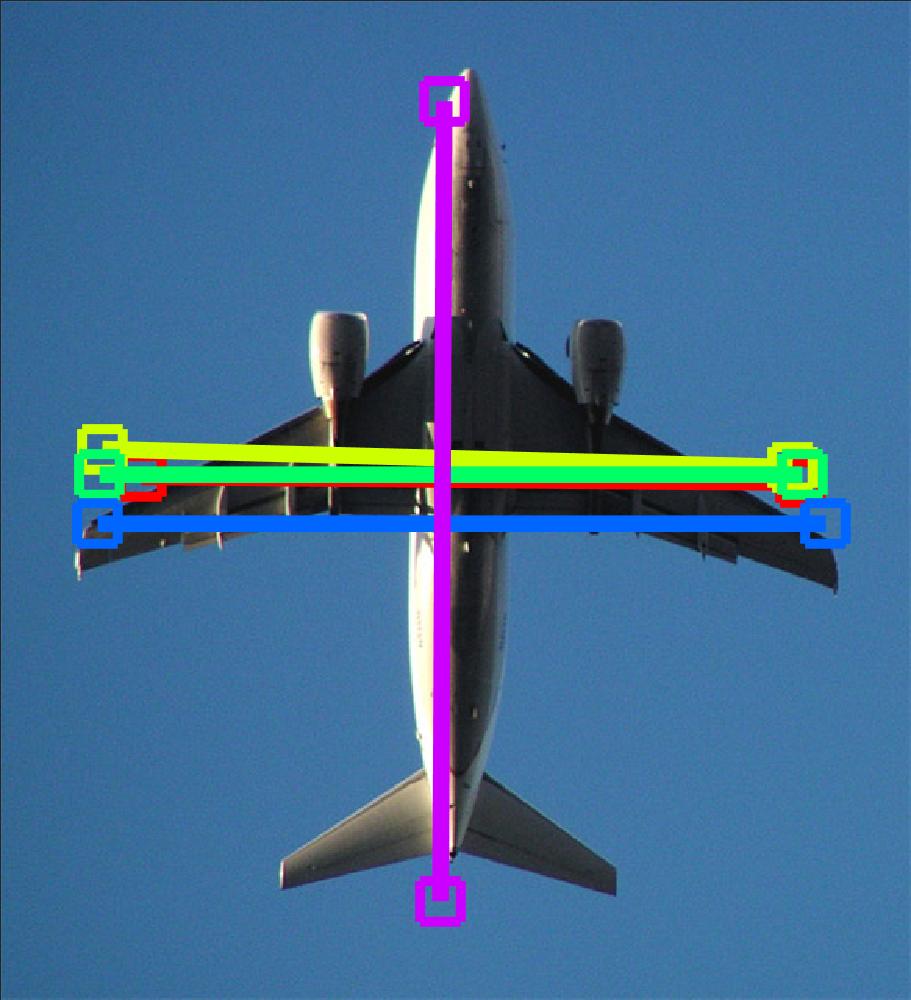}\label{fig:AVA1_Our} }
    \hspace{1mm}
    \subfloat[832486 - Lg]{\includegraphics[width=0.4\columnwidth]{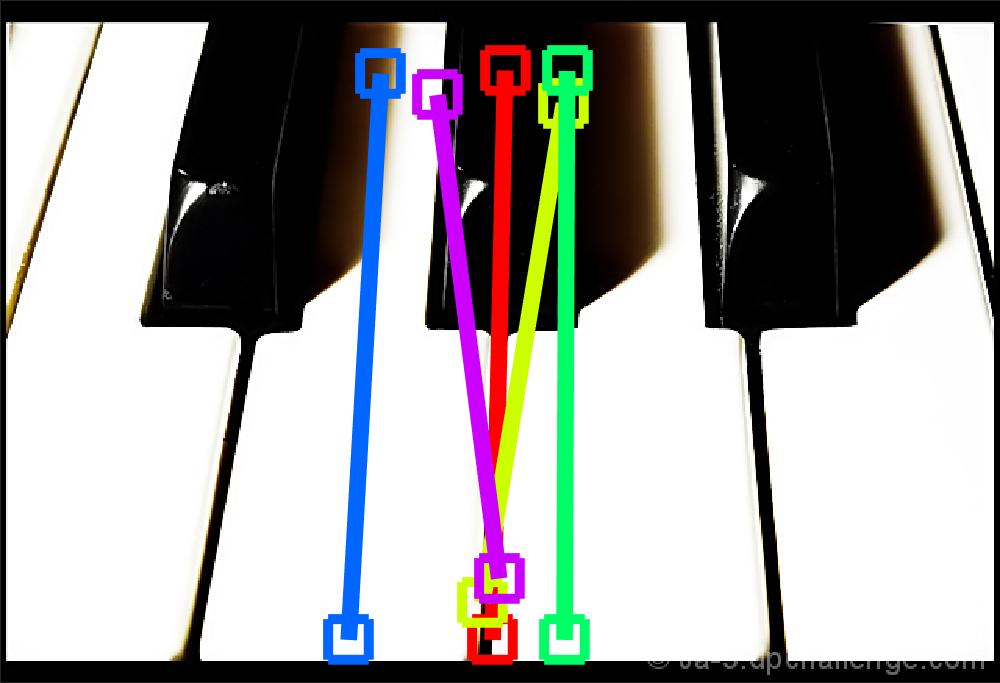}\label{fig:AVA2_Our} }
    \hspace{1mm}
    \subfloat[ref\_rm\_13 - Lg]{\includegraphics[width=0.39\columnwidth]{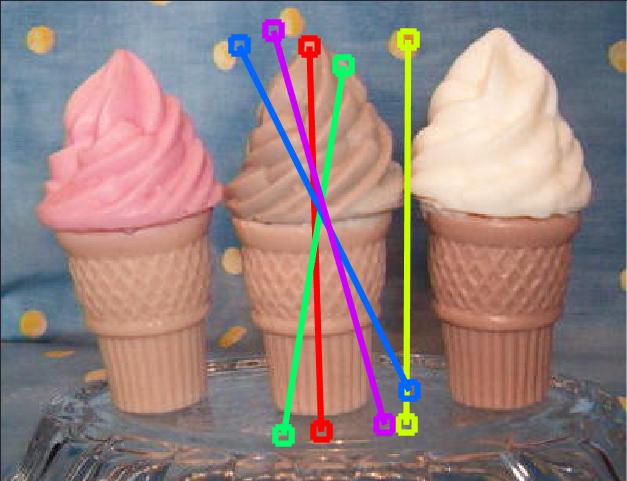}\label{fig:PSUm1_Our} } 
    \hspace{1mm}
    \subfloat[ref\_rm\_65 - Lg]{\includegraphics[width=0.4\columnwidth]{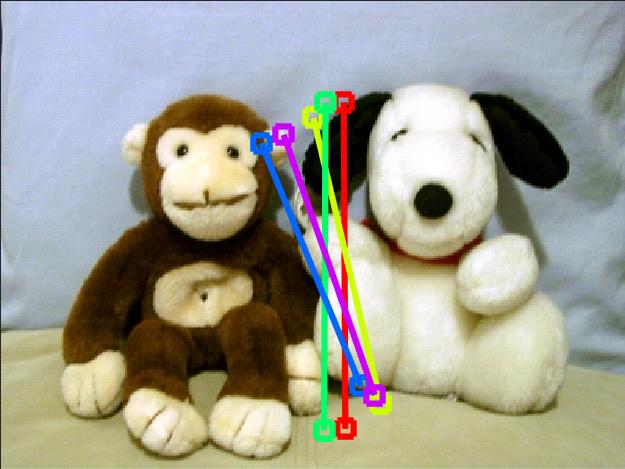}\label{fig:PSUm2_Our} } 
    \hspace{1mm}
    \subfloat[I034 - Lg]{\includegraphics[width=0.4\columnwidth]{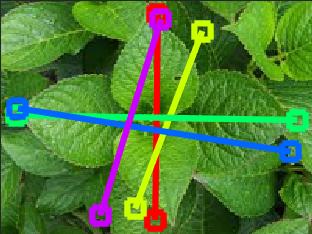}\label{fig:NYm1_Our} }
    \\
    \subfloat[16550 - Ela]{\includegraphics[width=0.32\columnwidth]{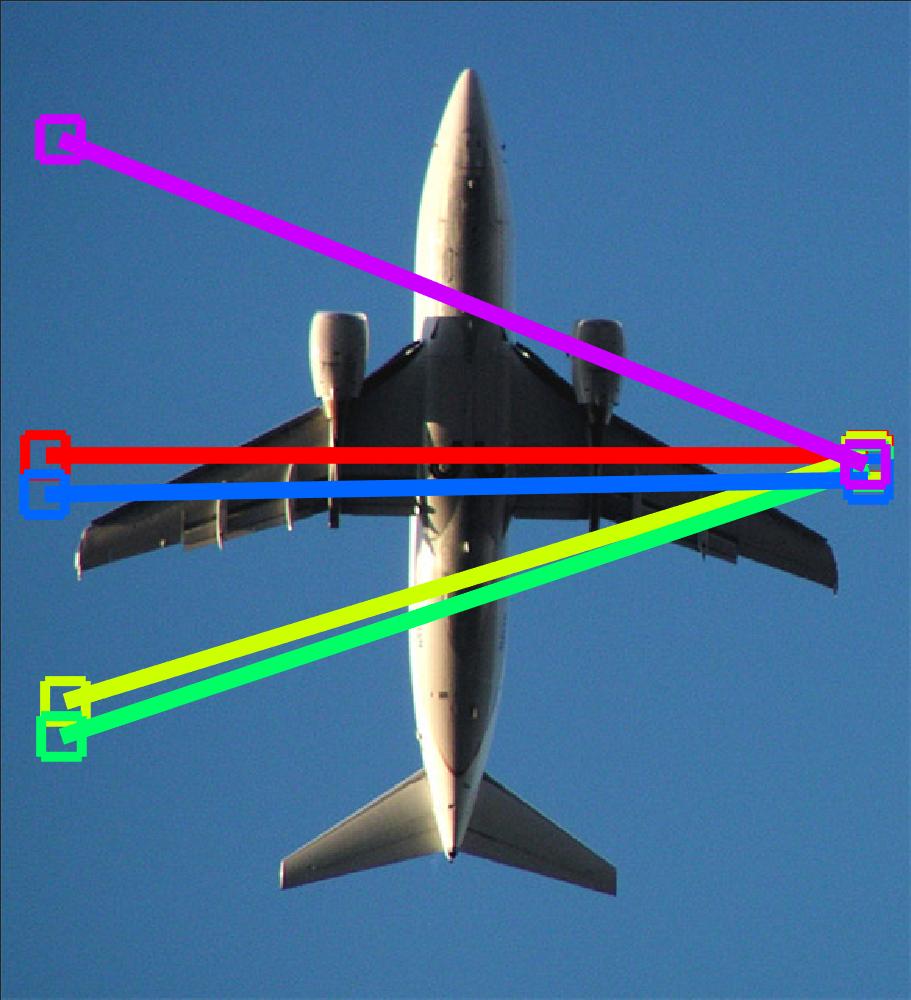}\label{fig:AVA1_Ela} }
    \hspace{1mm}
    \subfloat[832486 - Ela]{\includegraphics[width=0.4\columnwidth]{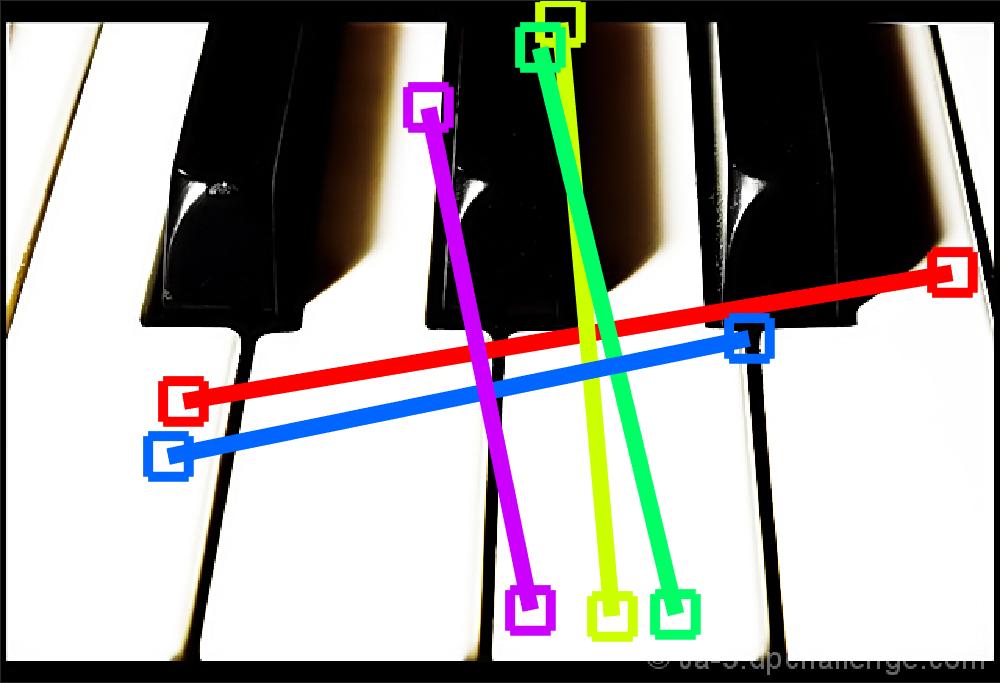}\label{fig:AVA2_Ela} }
    \hspace{1mm}
    \subfloat[ref\_rm\_13 - Ela]{\includegraphics[width=0.39\columnwidth]{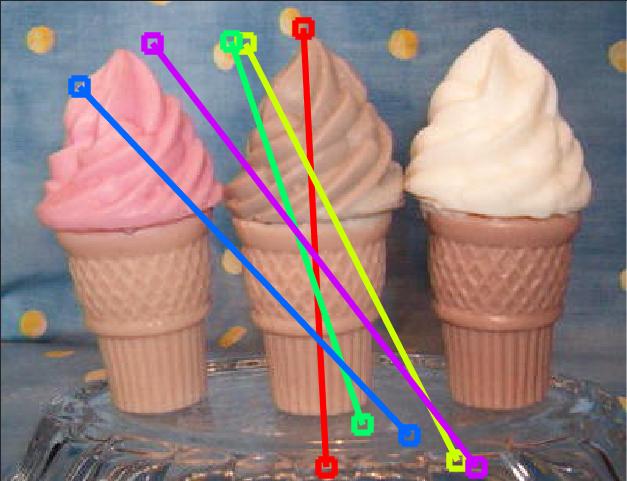}\label{fig:PSUm1_Ela} } 
    \hspace{1mm}
    \subfloat[ref\_rm\_65 - Ela]{\includegraphics[width=0.4\columnwidth]{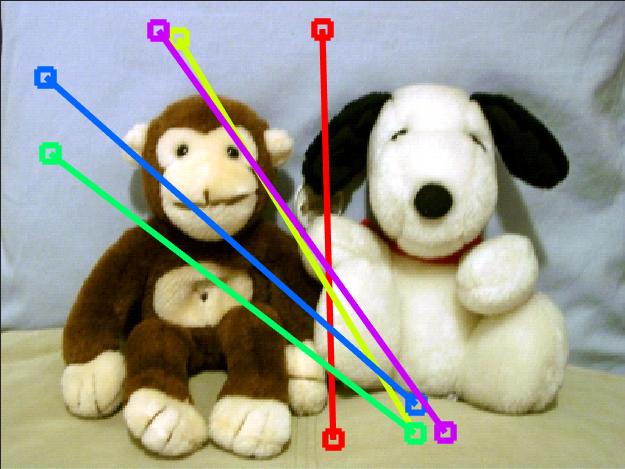}\label{fig:PSUm2_Ela} } 
    \hspace{1mm}
    \subfloat[I034 - Ela]{\includegraphics[width=0.4\columnwidth]{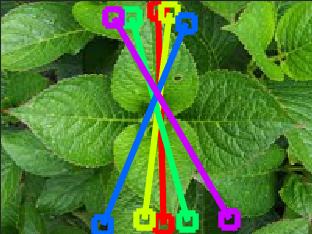}\label{fig:NYm1_Ela} }
    \\
    \subfloat[16550 - Loy]{\includegraphics[width=0.32\columnwidth]{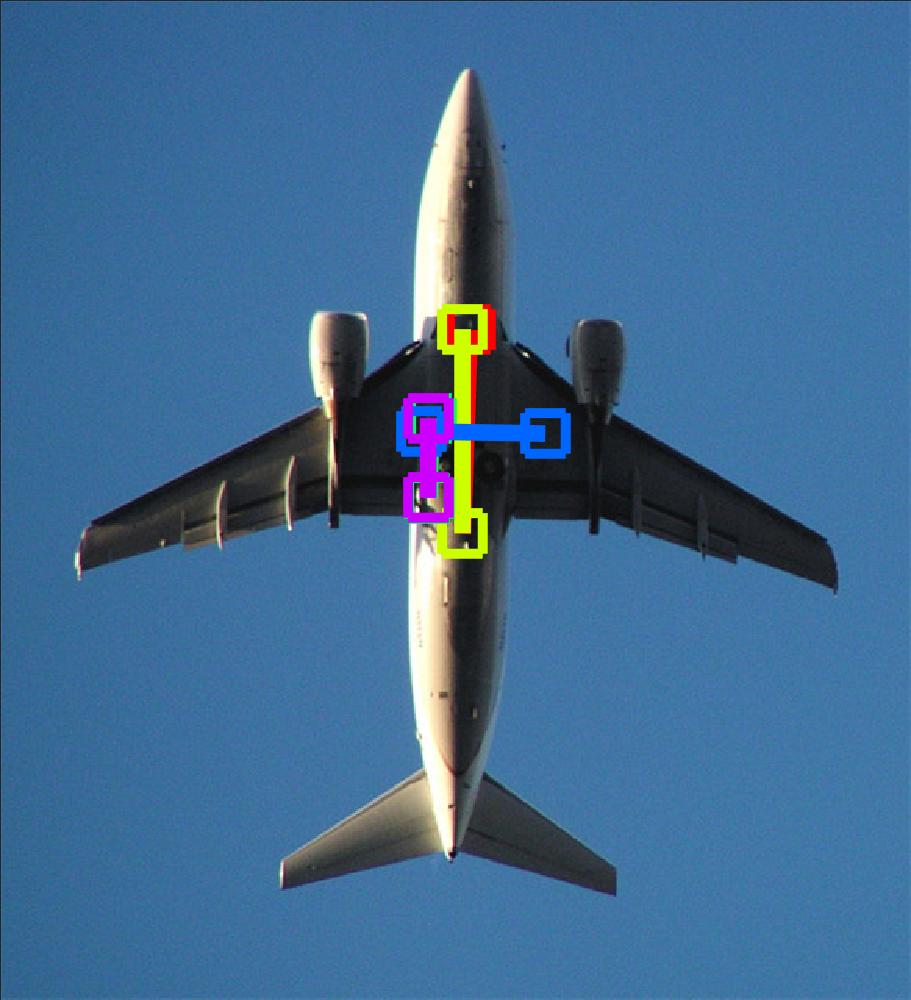}\label{fig:AVA1_Loy} }
    \hspace{1mm}
    \subfloat[832486 - Loy]{\includegraphics[width=0.4\columnwidth]{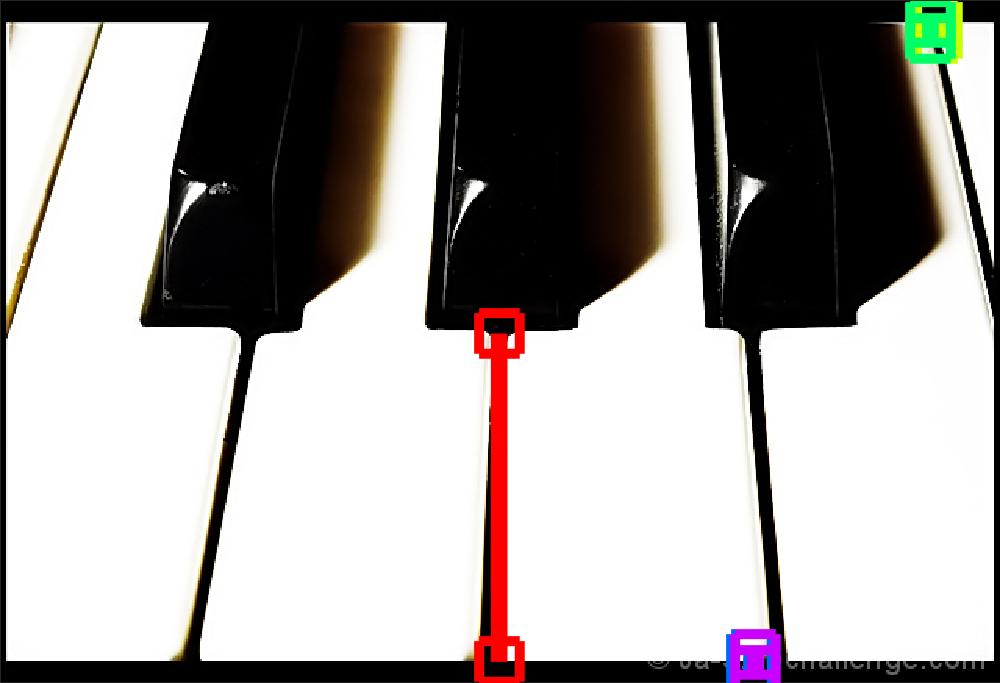}\label{fig:AVA2_Loy} }
    \hspace{1mm}
    \subfloat[ref\_rm\_13 - Loy]{\includegraphics[width=0.39\columnwidth]{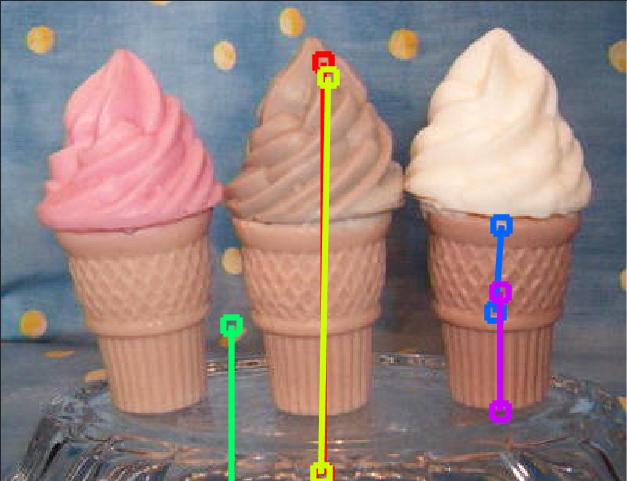}\label{fig:PSUm1_Loy} } 
    \hspace{1mm}
    \subfloat[ref\_rm\_65 - Loy]{\includegraphics[width=0.4\columnwidth]{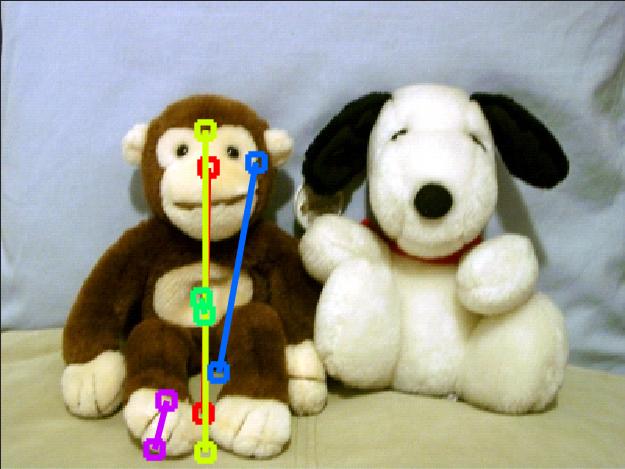}\label{fig:PSUm2_Loy} } 
    \hspace{1mm}
    \subfloat[I034 - Loy]{\includegraphics[width=0.4\columnwidth]{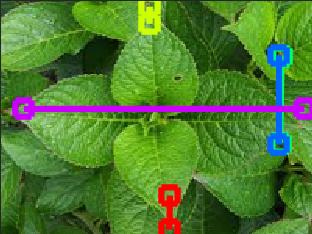}\label{fig:NYm1_Loy} }
	\caption{Some single-case and multiple-case results from AVA \cite{Elawady2016}, PSUm \cite{Rauschert2011,Liu2013}, and NYm \cite{Cicconet2016} datasets with groundtruth ({\color{blue} blue}) in 1st row (a-e). our methods in 2nd and 3rd rows (f-o) produces better results among El2016 \cite{Elawady2016} in 4rd row (p-t) and Loy2006 \cite{Loy2006} in 5th row (u-y). For each algorithm, the top five symmetry results is presented in such order: {\color{red} red}, {\color{yellow} yellow}, {\color{green} green}, {\color{blue} blue}, and {\color{magenta} magenta}. Each symmetry axis is shown in a straight line with squared endpoints. Best seen on screen.}
	\label{fig:res}
\end{figure*}

In our experimental evaluation, the algorithms are executed to detect and compare the global symmetries inside synthetic and real-world images. Tables~\ref{tab:resTP},~\ref{tab:resTPR17} show the different versions of true positive rates for the proposed methods (\textit{Lg} and \textit{LgC}) against Loy and Eklundh (\textit{Loy}) \cite{Loy2006}, Cicconet et al. (\textit{Cic}) \cite{Cicconet2014}, and Elawady et al. (\textit{Ela}) \cite{Elawady2016}. \textit{LgC} performs the best result among most cases in single and multiple symmetry, due to the importance of color information for the voting computations in colorful images. At the same time, \textit{Lg} has the top 2nd result, and sometimes the top 1st results in gray-scale or low-saturated images. \textit{Ela} \cite{Elawady2016} ranked as the top 3rd result, due to the utilization of small grids to compute window-based features. Thanks for the advantage of SIFT features, \textit{Loy} \cite{Loy2006} is still strong competent to be ranked as the top 4th result in general. \textit{Cic} \cite{Cicconet2014} has the lowest performance.

Figure~\ref{fig:prCurve} presents performance results in terms of precision and recall curves for single-case and multiple-case symmetry datasets, plus values of the maximum $F_1$ score to measure the performance of the proposed algorithms (\textit{Lg2017} and \textit{LgHSV2017}) against Loy and Eklundh (\textit{Loy2006}) \cite{Loy2006}, Cicconet et al. (\textit{Cic2014}) \cite{Cicconet2014}, and Elawady et al. (\textit{Ela2016}) \cite{Elawady2016}. In single-case symmetry, our method \textit{Lg2017} outperforms the other concurrent algorithms (\textit{Loy2006}, \textit{Cic2014}, and \textit{Ela2016}) in the context of using only gray-scale version of involved images. Furthermore, color version of our method \textit{LgHSV2017} exploits slightly improvement over gray-scale one \textit{Lg2017}. On the other hand, Only \textit{LgHSV2017} has better precision performance among others in (\textit{PSUm} and \textit{NYm}) datasets, due to many local symmetry groundtruth presenting inside multiple-case symmetry.

As a summary of the previous quantitative evaluations, figure~\ref{fig:res} compares qualitatively top performing algorithms showing different examples of reflection symmetry detection. Despite the single-case images (1st and 2nd columns) have strong shadowing effect in foreground objects, the color version of the proposed method \textit{LgC} easily finds the correct symmetry axes as a first candidate. On the other hand, the non-color proposed method \textit{Lg} satisfies the single-symmetry groundtruth in the two examples as first and fifth detections respectively. In contrast, \textit{Ela} \cite{Elawady2016} mismatches the provided groundtruth with all detection and \textit{Loy} \cite{Loy2006} only finds the local symmetry axes in the object parts having same contrast values. In multiple-case images (3rd, 4th and 5th columns), \textit{LgC} clearly detects the global and most of local symmetries. In the opposite side, the other methods struggle determining only more than one correct groundtruth. 
%%%%%%%%%%%%%%%%%%%%%%%%%%%%%%%%%%%%%%%%%%%%%%%%%%%%%%%%%%%%%%%%%%%%%%%%%%%%%%%%%
\section{Conclusion}
In this paper, we detect global symmetry axes inside an image using the edge characteristics of Log-Gabor wavelet response. For the purpose of improving our results, we additionally use textural and color histograms as local symmetrical measure around edge features. We show that the proposed methods provide a great improvement over single and multiple symmetry cases in all public datasets. Future work will focus on improving the voting representations, respect to the number of symmetry candidate pairs, resulting a precise selection of symmetrical axis peaks and their corresponding voting features. For art and aesthetic ranking applications, a unified balance measure is required to describe the global symmetry degree inside an image.
%%%%%%%%%%%%%%%%%%%%%%%%%%%%%%%%%%%%%%%%%%%%%%%%%%%%%%%%%%%%%%%%%%%%%%%%%%%%%%%%%
% \section*{Acknowledgements}
% We would like to thank Peter Kovesi for making the source code of Log-Gabor wavelet filters available on-line. We also thank Marcelo Cicconet and Gareth Loy to provide the MATLAB implementations of the reflection symmetry detection.
%%%%%%%%%%%%%%%%%%%%%%%%%%%%%%%%%%%%%%%%%%%%%%%%%%%%%%%%%%%%%%%%%%%%%%%%%%%%%%%%%
{\small
\bibliographystyle{ieee}

}
%%%%%%%%%%%%%%%%%%%%%%%%%%%%%%%%%%%%%%%%%%%%%%%%%%%%%%%%%%%%%%%%%%%%%%%%%%%%%%%%%
\end{document}